\documentclass[final,5p,times,authoryear,twocolumn]{elsarticle}
\usepackage[colorlinks, 
            linkcolor=blue,  
            anchorcolor=blue, 
            citecolor=blue,
            ]{hyperref}
\usepackage{graphicx}
\usepackage{subfigure}
\journal{NEURAL NETWORKS}
\usepackage{booktabs}
\usepackage{caption}
\usepackage{comment}
\usepackage{caption}
\usepackage{titlesec}
\titleformat{\subsection}
{\normalfont\bfseries}{\textbf{\thesubsection}}{1em}{\textbf}
\titleformat{\subsubsection}
{\normalfont\bfseries}{\textbf{\thesubsubsection}}{1em}{\textbf}
\usepackage{float}
\usepackage{afterpage}
\usepackage{amsmath}
\usepackage[T1]{fontenc}

\begin{document}
\begin{frontmatter}

\title{Color and Texture Dual Pipeline Lightweight Style Transfer}
\author{ShiQi Jiang,JunJie Kang, YuJian Li\affiliation{organization={Guilin University of Electronic Technology},
            city={Guilin},
            country={China}}}

\begin{abstract}
Style transfer methods typically generate a single stylized output of color and texture coupling for reference styles, and color transfer schemes may introduce distortion or artifacts when processing reference images with duplicate textures. To solve the problem, we propose a Color and Texture Dual Pipeline Lightweight Style Transfer (\textbf{CTDP}) method, which employs a dual pipeline method to simultaneously output the results of color and texture transfer. Furthermore, we designed a masked total variation loss to suppress artifacts and small texture representations in color transfer results without affecting the semantic part of the content. More importantly, we are able to add texture structures with controllable intensity to color transfer results for the first time. Finally, we conducted feature visualization analysis on the texture generation mechanism of the framework and found that smoothing the input image can almost completely eliminate this texture structure. In comparative experiments, the color and texture transfer results generated by \textbf{CTDP} both achieve state-of-the-art performance. Additionally, the weight of the color transfer branch model size is as low as 20k, which is 100-1500 times smaller than that of other state-of-the-art models.
\end{abstract}
\afterpage{
	\begin{figure*}[t]
		\centering
		\includegraphics[width=\textwidth,height=\textheight,keepaspectratio]{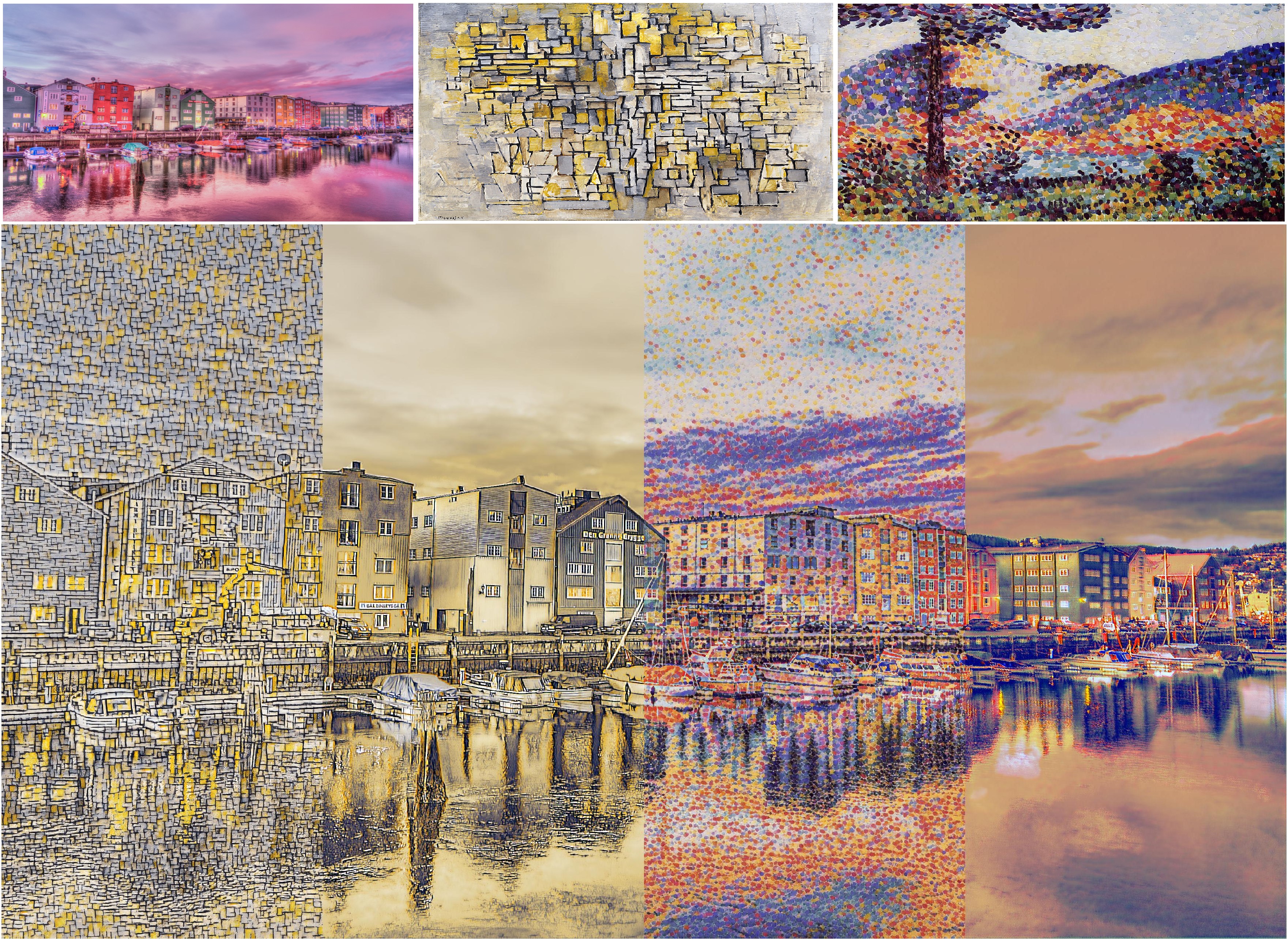}
		\caption{\textbf{The 4K super-resolution stylized image} generated by our proposed CTDP. The top of the image displays a content image and two style images, with each style image generating a pair of stylized results  (colors, textures) that are concatenated together.}
		\label{4ktupian}
	\end{figure*}
}

\end{frontmatter}

\section{Introduction}
Style transfer is a highly attractive image processing technique that can transfer the unique colors and texture styles of artworks to content images. In recent years, methods for style transfer have been widely proposed, which can be roughly divided into two categories: online image optimization and model optimization.

The representative of image optimization methods is (\cite{gatys}), which innovatively transfers gradients to the input image and iteratively optimizes the input content image directly. The style pattern is represented by the feature correlation of deep convolutional neural networks (VGG, \cite{vgg}). Subsequent work mainly focuses on different forms of loss functions (\cite{15,27}). However, this slow online optimization method has a high time cost and greatly reduces its actual citation value. In contrast, the model optimization method effectively solves the time-consuming problem of online iteration through offline model training and forward reasoning. There are three main types of model optimization: (1) Training exclusive style transformation models for a single artistic style (\cite{Perceptual,18,32,33}) Synthesize stylized images using a single given artistic style image; (2) Training model that can convert multiple styles (\cite{3,7,35,20,37}) Introducing various network architectures while handling multiple styles; (3) Arbitrary style transformation model (\cite{12,21,microast,collaborative,meta,DynamicIN}) used different mechanisms such as feature modulation and matching to transfer any artistic style.

Looking back at all the above methods, we found that most models for style images simply match the Gram (\cite{gatys}) or other statistics output in the middle of VGG. However, style images are composed of color and texture structures, which not only have rich colors but also highly repetitive texture structure information. We hope to decouple and match the color and texture structure information of style images. In fact, due to the inexplicability of the VGG intermediate layer, the decoupling of color and texture information in the Gram matrix calculated from its intermediate feature map is even more challenging, as it couples multi-dimensional and very complex information such as color, texture, and semantics. Therefore, despite significant progress in recent years, existing methods still cannot decouple the matching of color and texture structure information.

\begin{figure*}[t]\centering
	\centering
	\includegraphics[width=\textwidth,height=\textheight,keepaspectratio]{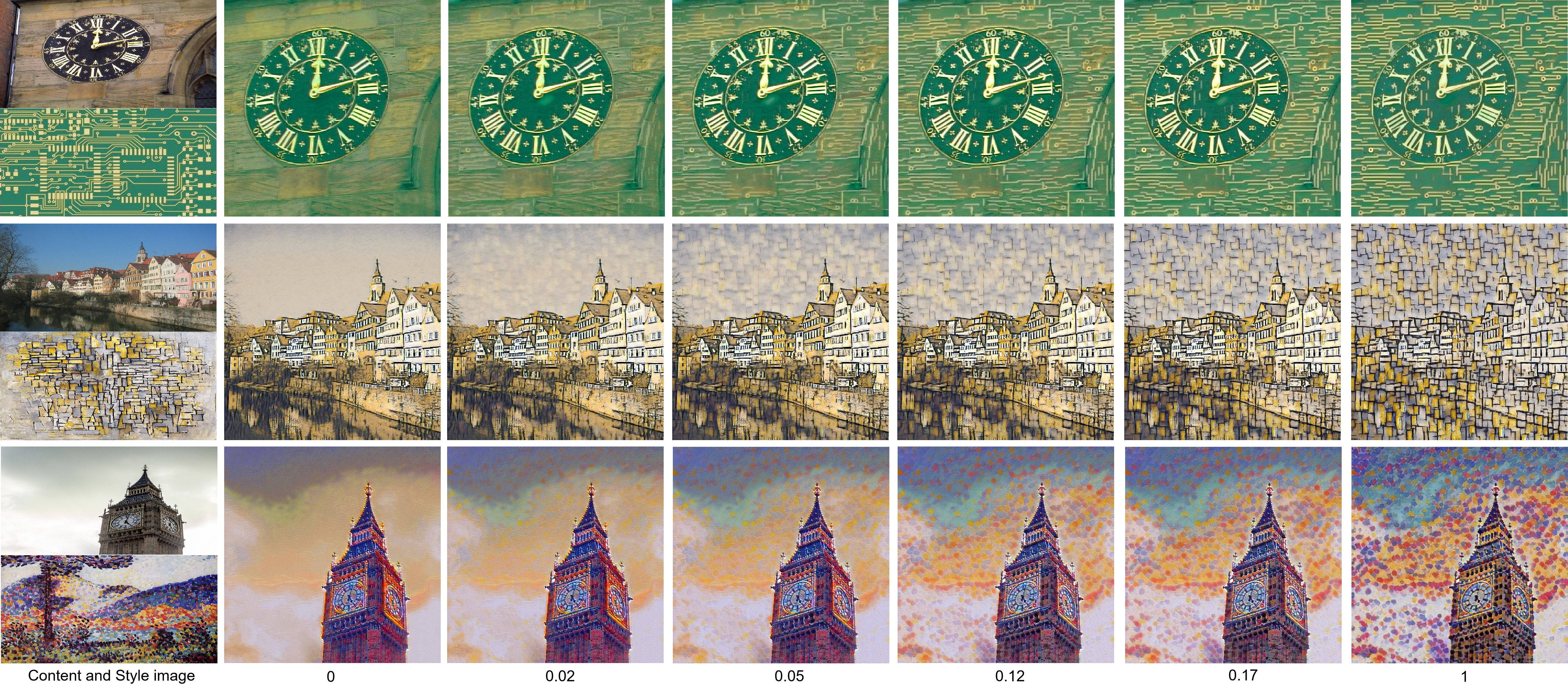}
	\caption{Add a texture structure with controllable intensity to the color transfer result, and the following values represent the intensity modulation parameter $\lambda_{d}$ of the texture feature.}
	\label{inter}
\end{figure*}

\begin{figure*}[t]
	\centering
	\includegraphics[width=\textwidth,height=\textheight,keepaspectratio]{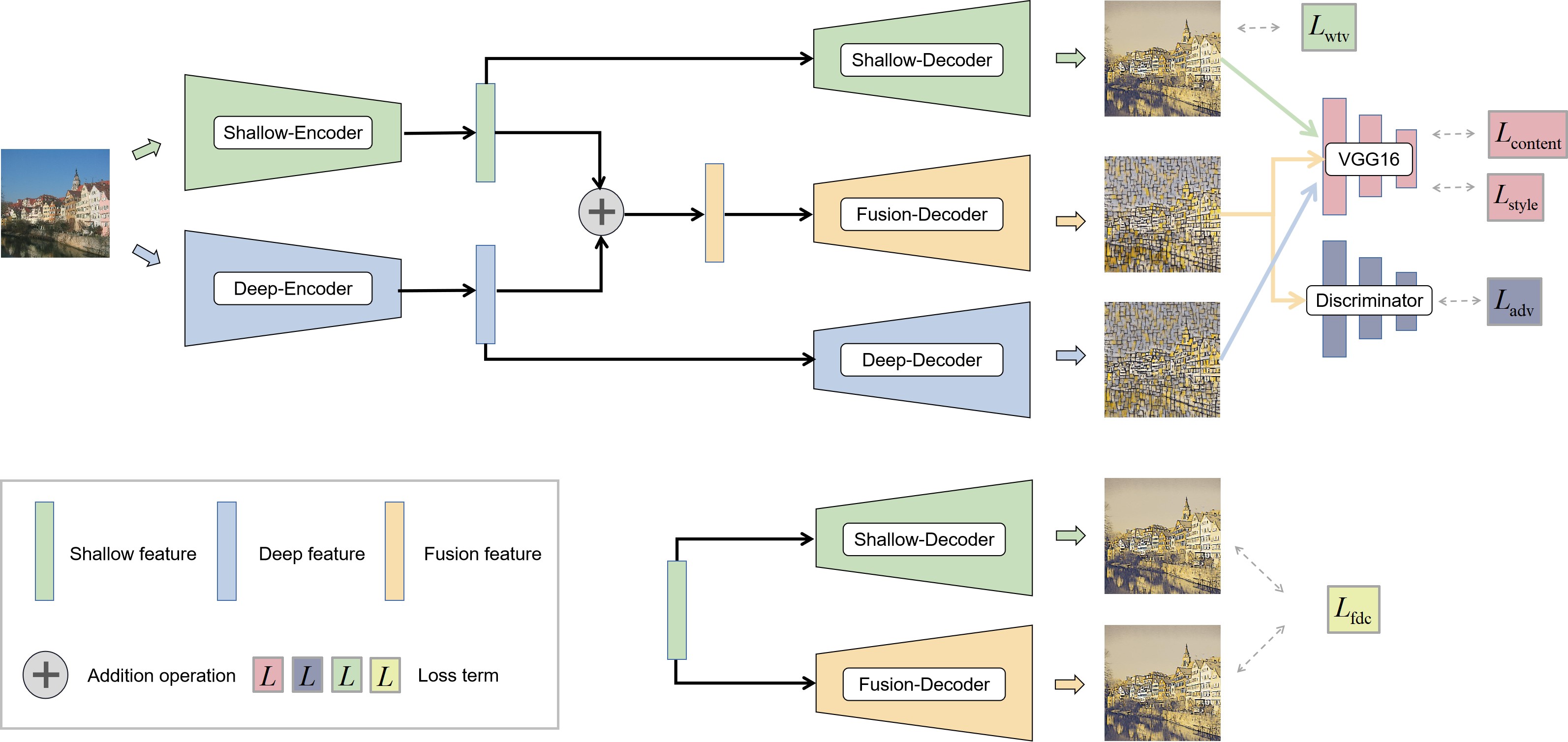}
	\caption{Architecture illustration of the proposed \textbf{CTDP}. See Section 3 for details.}
	\label{jiegou}
\end{figure*}

In the face of the aforementioned challenges, we propose a dual pipeline model that can quickly generate both color and texture transfer results simultaneously. Inspired by the DcDae (\cite{DcDae}) method, which directly decodes shallow features and outputs them, but this is only a side effect of model training. If it is directly treated as a color transfer result, it often produces completely incorrect results, as shown in Fig.\ref{duibi}. Our key insight is to use a dual pipeline framework, using different VGG level Gram matrices for loss calculation for color and texture transfer branches. In the color transfer branch, we do not directly decouple the color information in Gram, but instead suppress the texture feature representation through our designed masked total difference loss.

In addition to the above improvements, we are able to add texture structures with controllable intensity to color transfer results for the first time, which is completely different from DcDae (\cite{DcDae}) adding texture information from a similar color transfer result. Compared with state-of-the-art models, our CTDP can achieve better color and texture transfer effects than existing methods. In summary, our contributions are as follows:
\begin{itemize}
    \item We propose a lightweight dual pipeline framework that can quickly generate both color and texture transfer results simultaneously.
    \item Propose a masked total variation loss to suppress artifacts and texture structure representation in color transfer branches without affecting the semantic part of the content.
    \item For the first time, we are able to add texture structures with controllable intensity to color transfer results.
    \item Detailed feature visualization analysis of texture generation mechanism and found that input smoothing operation can almost completely eliminate texture structure representation.
    \item Numerous qualitative and quantitative experiments have shown that our method can quickly achieve high-quality color and texture style transfer simultaneously.
\end{itemize} 

\section{Related work}
\subsection{Neural Style Transfer}
With the groundbreaking work of (\cite{gatys}), the era of neural style transfer (NST) has arrived. The visual appeal of style transfer has inspired subsequent researchers to improve in many aspects, including efficiency (\cite{Perceptual,32}); Quality (\cite{stroke,17,10,xie,DcDae}); Diversity (\cite{wang,chen}) and User Control (\cite{zhang,cham}); Despite significant progress, existing methods still cannot decouple the color and texture structure information of matching style images, nor can they simultaneously complete the two tasks of color and texture transfer.

\subsection{Color Style Transfer}
Unlike artistic style transfer (\cite{stroke,17,10,xie,meta,microast,demystifying,SANet}), it usually changes both color and texture structure simultaneously. The purpose of color style transfer (also known as realistic style transfer) is to only transfer colors from one image to another. Traditional methods (\cite{pitie2005n,pitie2007automated,reinhard2001color}) mostly match statistical data of low-level features, such as the mean and variance of images (\cite{reinhard2001color}) or histograms of filter responses (\cite{pitie2005n}). However, if there is a significant appearance difference between the style and the input image, these methods typically transfer unwanted colors. In recent years, many methods for color transfer using convolutional deep learning methods (\cite{photowct2,closed,deep,photorealistic,cap}) have been proposed. For example, (\cite{photorealistic}) Introduced a model with wavelet pooling to reduce distortion. CAP-VSTNet (\cite{cap}) uses a reversible residual network and an unbiased linear transformation module to prevent artifacts. Previous methods have improved in suppressing artifacts and content preservation, but have overlooked the impact of complex textures in reference styles on color transfer. The proposed method solves this problem by reducing receptive fields and masked total variation loss to suppress texture representation in Gram (\cite{gatys}).

\section{Method}
Given an arbitrary content image, our goal is to generate both color and texture transfer images simultaneously, and we are able to add texture structures with controllable intensity to color transfer results. The challenges of this task mainly lie in four aspects: (1) The shallow feature output in DcDae (\cite{DcDae}) is not a result of color transfer but a side effect of model training, and cannot be directly used for color transfer tasks; (2) In color transfer tasks, using Gram matrices to match color information inevitably introduces a large amount of texture structural information. How to decouple or suppress these structural information; (3) How to ensure that color transfer branch feature maps can also decode correct color transfer results in a fusion decoder; (4) This method should be able to generate high-quality color and texture transfer results for any content image simultaneously.

\subsection{Overview of CTDP}
As shown in Fig.\ref{jiegou}, our CTDP framework consists of four main components: shallow encoder $Enc_s$ decoder $Dec_s$, deep encoder $Enc_d$ decoder $Dec_d$, fusion decoder $Dec_f$, and style discriminator $D_s$ (only used during the training phase). Shallow layers are responsible for color transfer tasks, while deep layers are responsible for texture transfer tasks. The fusion decoder is responsible for outputting the results of the fusion of shallow and deep features. Paired encoders and decoders have symmetric lightweight structures, consisting of the first two standard convolutional layers and several deep separable convolutional layers (DW, \cite{mobilenets}) in the middle.

Specifically, the shallow layer is a straight tube structure without stride-2 convolution, while the deep layer has two stride-2 convolutions. Shallow features are fused with deep features after detail attention-enhanced ($Dae$, \cite{DcDae}) and stride-2 convolution. The forward inference pipeline of our framework is as follows:

(1) Extracting the shallow features $f_s$ of content image $C$ using a shallow encoder $Enc_s$, denoted as $f_s := Enc_s(C)$.

(2) Extracting the deep features $f_d$ of content image $C$ using a deep encoder $Enc_d$, denoted as $f_d := Enc_d(C)$.

(3) Obtain color transfer output $CS_ c$ by inputting shallow features $f_s$ into shallow decoder $Dec_s$, denoted as $CS_ c := Dec_s(f_s)$.
 
(4) Obtain texture transfer output $CS_ t$ by inputting shallow features $f_d$ into shallow decoder $Dec_d$, denoted as $CS_ t := Dec_d(f_d)$. 

(5) Obtain fusion features $f_ f$ by adding shallow features $f_s$ with $Dae$ to deep features $f_d$, denoted as $f_ f := \lambda_{s}Dae(f_s) + \lambda_{d}f_d$, where $\lambda_{s}$ and $\lambda_{d}$ represent the fusion strength of shallow and deep features, respectively. The $Dae$ here contains a stride-2 convolution.

(6) Obtain fusion features transfer output $CS_ f$ by inputting fusion features $f_ f$ into fusion decoder $Dec_f$, denoted as $CS_ f := Dec_f(f_f)$.

\textbf{Training Losses}. In order to achieve style transfer, similar to the previous method (\cite{gatys,xie,meta,microast,demystifying,SANet,adain,DcDae}), we use pre trained VGG-16 (\cite{vgg}) as our loss model to calculate content and style loss. We use perceptual loss (\cite{Perceptual}) as our branch content loss $\mathcal{L}_{bc}$, and all three of our branch content losses are calculated in the ${relu2\_1}$ layers of VGG-16. The branch style loss $\mathcal{L}_{bs}$ is defined as the matching Gram matrix (\cite{gatys}), and the three branches calculate the style loss at different levels (see details in Sec.\ref{sec:bs}). Introduce style discrimination loss $\mathcal{L}_{adv}$ similar to (\cite{DcDae}) to ensure the overall color and texture matching effect of stylized images. Please note that we only use VGG-16 during the training phase and do not require complex loss calculations or involve any large networks during the inference phase.

To further suppress texture representation, we designed a masked total variation loss $\mathcal{L}_{mtv}$ to suppress texture representation in content smoothing regions (see details in Sec.\ref{sec:mtv}). Additional $\mathcal{L}_{fdc}$ loss was designed  to ensure consistency in decoding shallow features between shallow decoder and fusion decoder (see details in Sec.\ref{sec:fdc}). In summary, the overall goals of our CTDP are:
\begin{equation}
	\label{zongloss}
	\mathcal{L}_{Full} = \lambda_{bc}\mathcal{L}_{bc} + \lambda_{bs}\mathcal{L}_{bs} + \lambda_{adv}\mathcal{L}_{adv} + \lambda_{mtv}\mathcal{L}_{mtv} + \lambda_{fdc}\mathcal{L}_{fdc},
\end{equation}
where hyper-parameters $\lambda_{bc}$, $\lambda_{bs}$, $\lambda_{adv}$, $\lambda_{mtv}$ and $\lambda_{fdc}$ define the relative importance of each component in the total loss function.

\subsection{Dual Pipeline}
The dual pipeline framework refers to our model outputting both color transfer and texture transfer results simultaneously. The model actually has three decoder branches, which decode shallow, deep and fusion features respectively. We generally specify the output of the shallow decoder as the color transfer result, and the output of the fusion decoder as the texture transfer result.
\subsubsection{Branch Style Loss}
\label{sec:bs}
In order to constrain the three branches for different stylization tasks, we apply Gram matrix constraints at different levels of VGG to the three stylized outputs. (1) For the shallow branch responsible for color transfer, matching the color information from the shallower layers of VGG is sufficient. Therefore, we only calculate the Gram matching loss at layers \{${relu1\_1},{relu2\_1}$\}. (2) For the deep branch dedicated to texture transfer, matching the repetitive texture structure features from the deeper layers of VGG is necessary. Hence, we compute the Gram matching loss at layers \{${relu2\_1},{relu3\_1},{relu4\_1}$\}. (3) For the fusion branch, which aims to better integrate features from both shallow and deep layers, we directly calculate the Gram matching loss at layers \{${relu1\_1},{relu2\_1},{relu3\_1},{relu4\_1}$\}. The branch style loss $\mathcal{L}_{bs}$ is the sum of the losses of each of the three branches.

As shown in Fig.\ref{duibi}, treating the shallow feature output of DcDae as a color transfer result would be very poor. This is because its shallow feature output has no constraints and is just a side effect of the style transfer task. In contrast, our CTDP achieves direct constraints on shallow decoding through branch style loss, which is the first time that the output of shallow feature decoding can be directly used for color transfer tasks, as shown in Fig.\ref{duibi}.

\begin{figure}[t]
	\centering
	\subfigure[Shallow decoder output]{\includegraphics[width=0.156\textwidth]{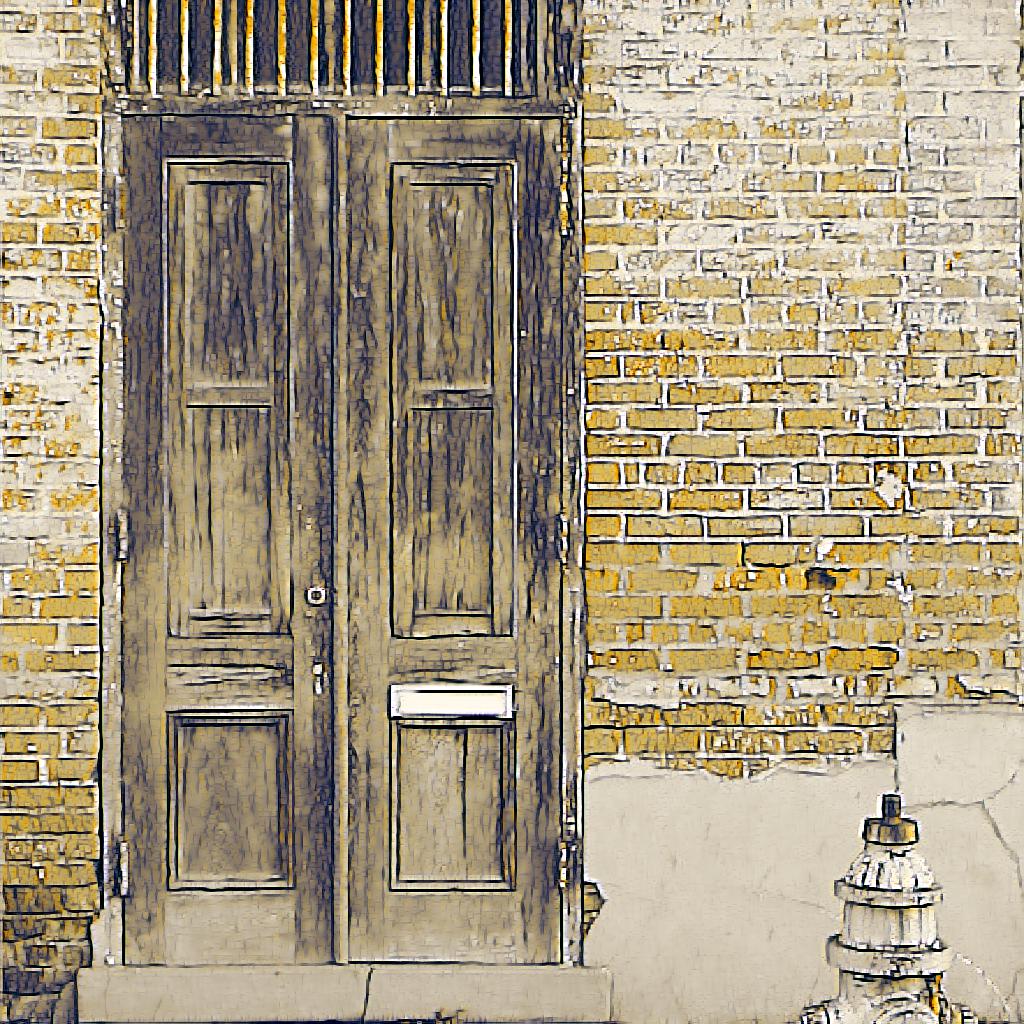}}
	\subfigure[Fusion decoder output (w/o $\mathcal{L}_{fdc}$ )]{\includegraphics[width=0.156\textwidth]{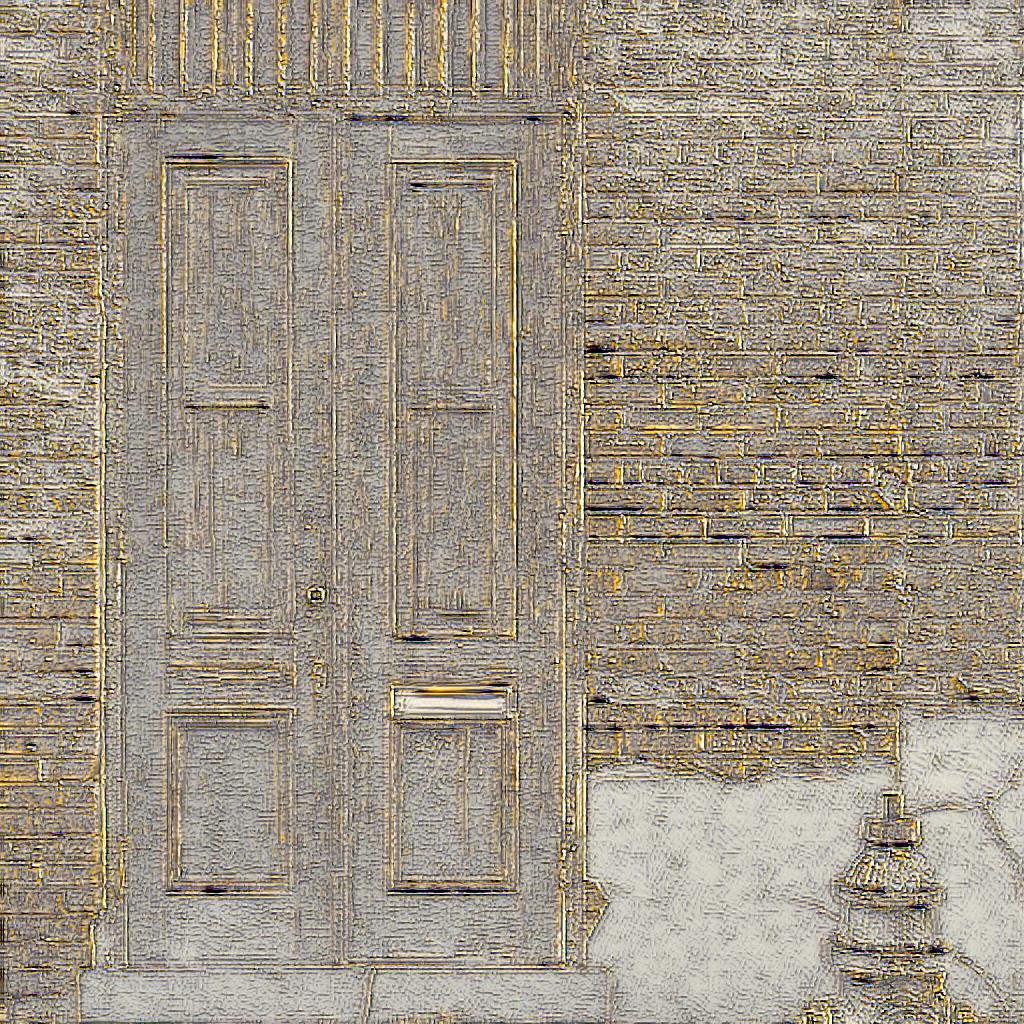}}
	\subfigure[Fusion decoder output (w/ $\mathcal{L}_{fdc}$ )]{\includegraphics[width=0.156\textwidth]{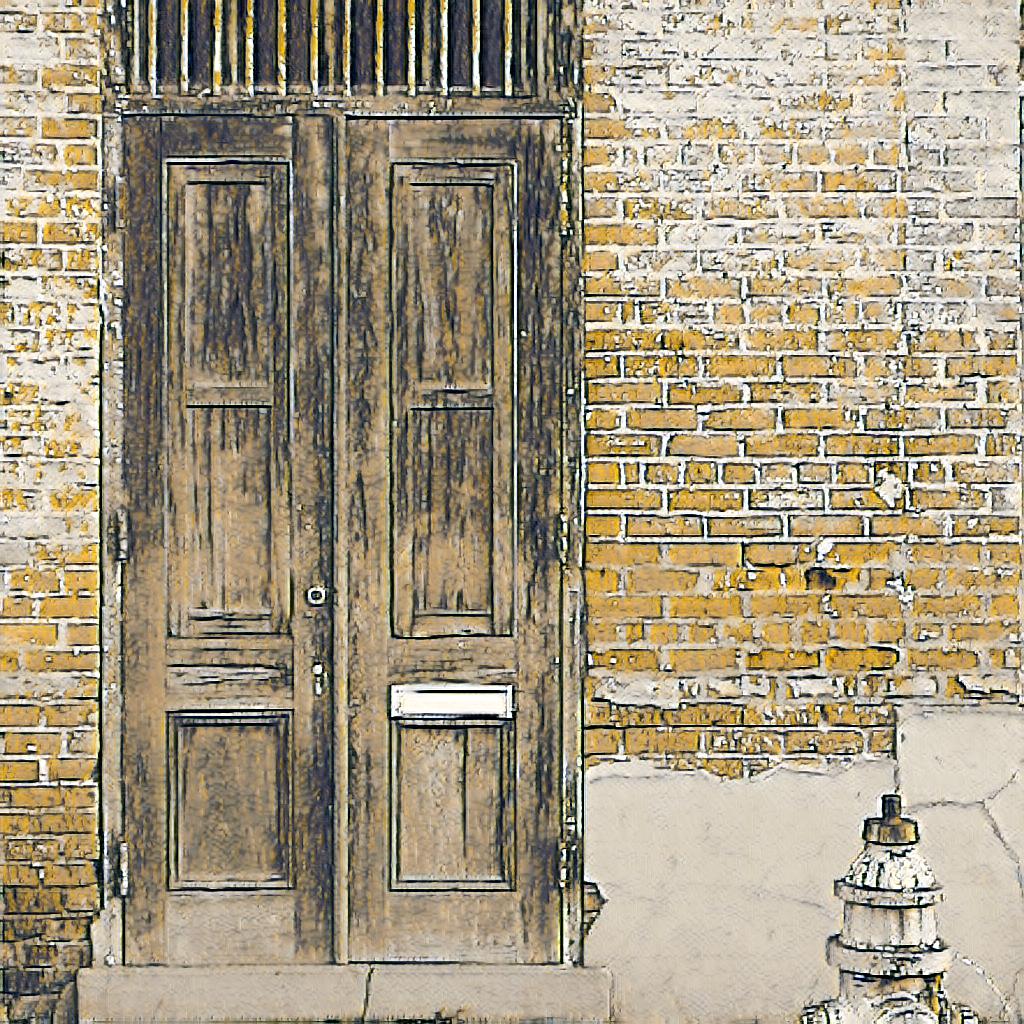}}
	\caption{\textbf{Ablation study of feature decoding consistency loss.} $\mathcal{L}_{mtv}$ ensures that shallow features in the shallow and fusion decoders' outputs yield similar results.}
	\label{fdc}
\end{figure}

\subsubsection{Feature Decoding Consistency}
\label{sec:fdc}
Although the shallow feature ${f_s}$ is trained using the loss of ${relu1\_1}$ and ${relu2\_1}$ Gram matrices in the color transfer branch, if the ${f_s}$  are directly input to the fusion decoder $Dec_f$ for output, an error as shown in Fig.\ref{fdc}(b) will occur. This may be due to the lack of loss constraints on the direct output of shallow features by the fusion decoder, which, like DcDae (\cite{DcDae}), is only a side effect product of color transfer tasks. So we propose feature decoding consistency loss $\mathcal{L}_{fdc}$ to ensure the decoding consistency of shallow features between shallow decoders and fusion decoders. 

In our dual pipeline framework, $c^{'}\in R^{B \times 3 \times H \times W}$ is a batch of input content images that are resized to $512$ and then randomly cropped to $256$, and $B$ is the size of a batch.The feature decoding consistency loss $\mathcal{L}_{fdc}$ is defined as:
\begin{equation}
	\label{fdcloss}
	\mathcal{L}_{fdc} = \frac{1}{3HW}\left\lVert Dec_f(Dae(Enc_s(c^{'}))) - Dec_s(Enc_s(c^{'}))\right\rVert_2^{2}.
\end{equation}
This loss is the Euclidean distance in pixel space between shallow features passed through the shallow and fusion decoders' outputs.
\subsubsection{Controllable Texture Intensity}
Unlike DcDae, we are able to add texture structures with controllable intensity to the color transfer results, as shown in Fig.\ref{inter}. When $\lambda_{d}$ is 0, it is equivalent to decoding the shallow feature map directly and outputting it, resulting in the color transfer result (DcDae's shallow feature direct output is a meaningless result). As the value of $\lambda_{d}$ continues to increase, the intensity of deep feature maps increases, and it can be observed that repeated texture structures in stylized images are becoming more apparent.

\begin{figure}[t]\centering
	\subfigure[w/ $\mathcal{L}_{bs}$]{\includegraphics[width=0.156\textwidth]{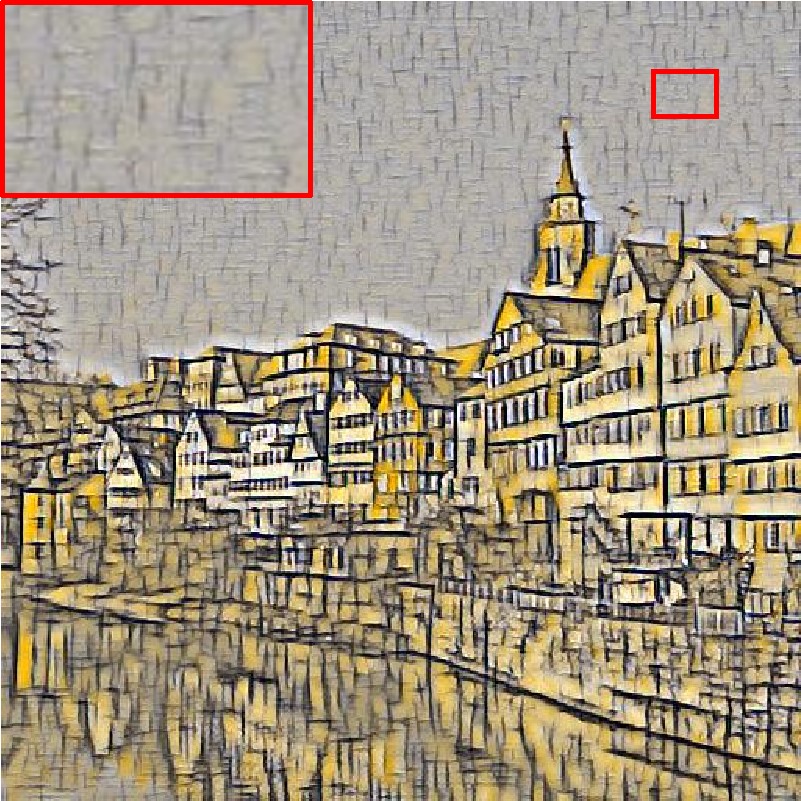}}
	\subfigure[w/o kernel 9 convolution and downsampling]{\includegraphics[width=0.156\textwidth]{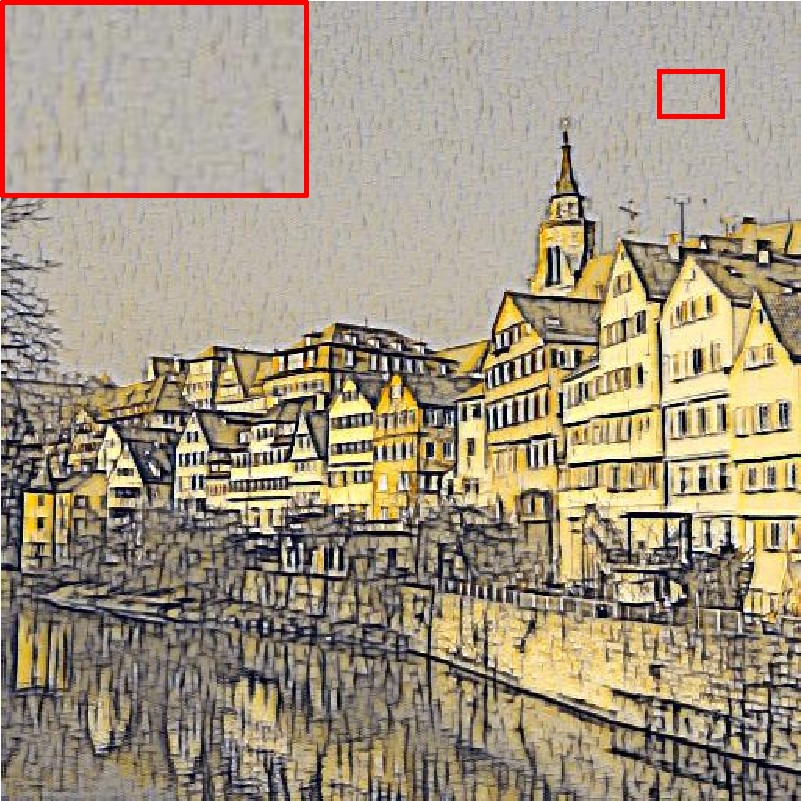}}
	\subfigure[w/ $\mathcal{L}_{mtv}$ ]{\includegraphics[width=0.156\textwidth]{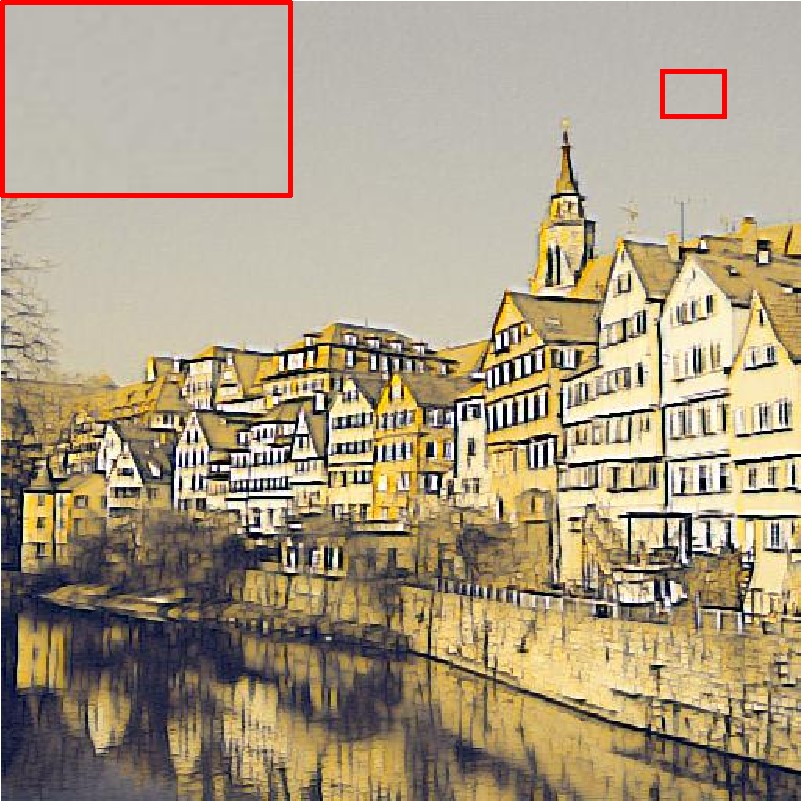}}
	\caption{Gradually suppress the texture by overlaying three methods.}
	\label{ts}
\end{figure}

\begin{figure*}[t]\centering
	\centering
	\includegraphics[width=\textwidth,height=\textheight,keepaspectratio]{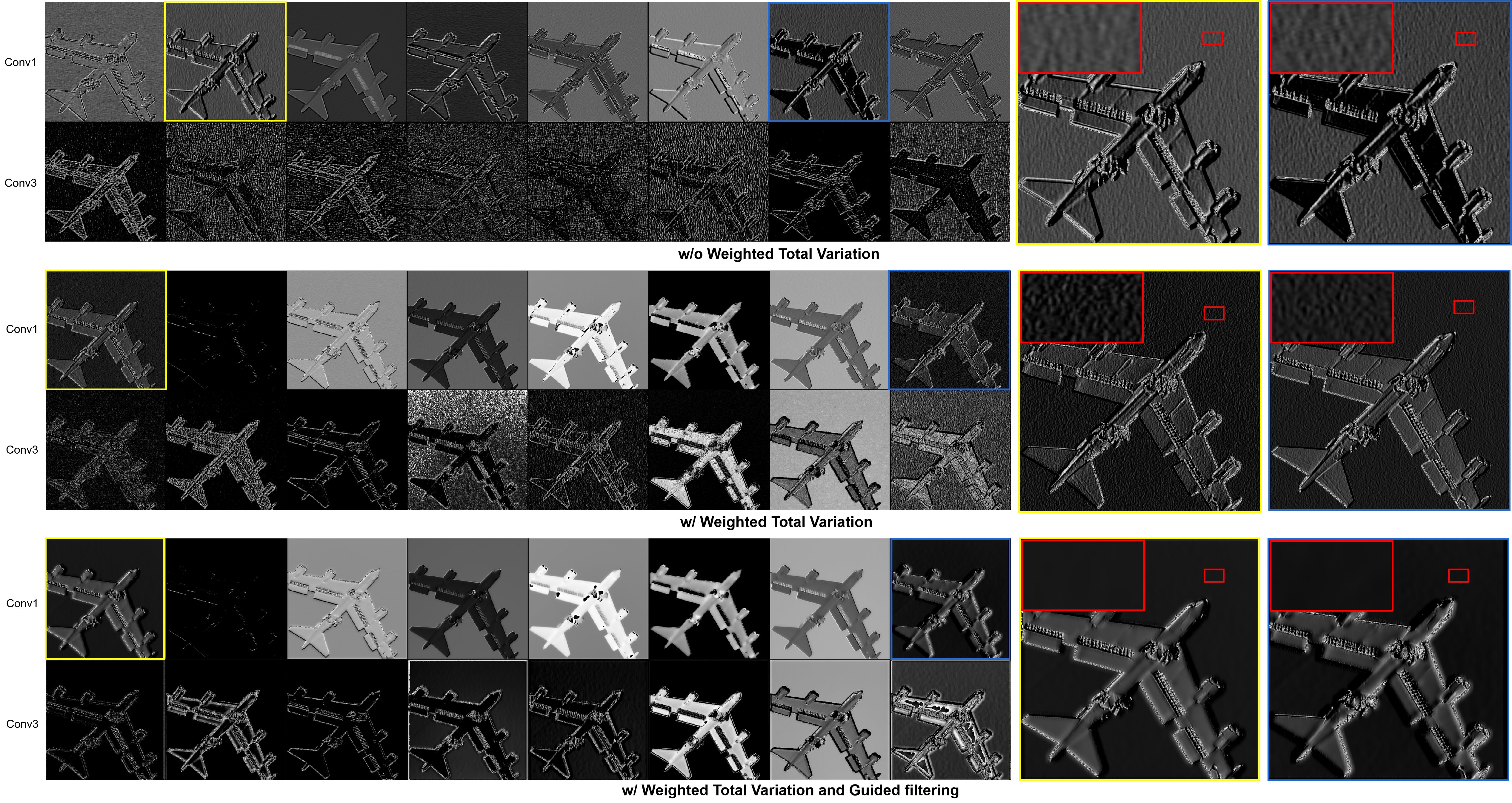}
	\caption{Visualization of feature maps for the first and third convolutions of three methods.}
	\label{vis}
\end{figure*}

\subsection{Texture Suppression}
The VGG model itself is inexplicable, and the Gram matrix calculated from its intermediate feature maps is even more inexplicable and highly coupled with various information. Therefore, it is extremely difficult to separate the color information in the Gram matrix through decoupling. We take another approach to achieve color transfer tasks by suppressing texture generation capabilities and outputting texture representations.

\subsubsection{Reduced Receptive Field}
\textbf{Style feature receptive field.} As described in Sec.\ref{sec:bs}, our color transfer branch only matches the Gram matrix of the first two layers of VGG output, which limits the receptive field of style features and allows the Gram matrix to encode only shallower style features. From the perspective of style feature receptive field, this approach reduces the high-level texture representation in the target style loss function.

\textbf{Shallow model receptive field.} As shown in Fig.\ref{ts}(a), simply using branch style loss is not enough. Although calculating only the Gram loss of two layers can indeed reduce many repetitive large color blocks and texture structures, subtle line structures still exist. We replaced the convolution with the first and last kernels of 9 with convolution kernel 3 and removed all downsampling methods to reduce the receptive field of the shallow model. As shown in Fig.\ref{ts}(b), the linear texture structure was further suppressed. From the perspective of the shallow model receptive field, this method suppresses the texture encoding and decoding ability of the shallow model.
\subsubsection{Masked Total Variation}
As shown in Fig.\ref{ts} (a-b), reducing style features and shallow model receptive fields alone cannot completely eliminate texture structure. We also need to add stronger constraints to the loss function.

\textbf{EdgeMask.} While suppressing texture, we hope that the edges and semantic information of the content image are not disturbed, so we hope to extract the edge information of the input content image as an mask to avoid suppressing the semantic information. The calculation process of $EdgeMask\in R^{B \times 1 \times H \times W}$ can be defined as:
\begin{equation}
	\label{mask}
	EdgeMask = threshold(Edge(c^{'}), \delta),
\end{equation}
where the $Edge$ represents the edge detection Sobel operator, $threshold$ represents a binary function that maps values less than $\delta$ to 0, and values greater than $\delta$ to 1 ($\delta$ is set to 0.2 by default).

After the above steps, we will obtain a binary $EdgeMask$ with the same size as $c^{'}$, which corresponds to an edge information region value of 1 in the content image and a smooth background region value of 0. Afterwards,  $EdgeMask$ will act as a mask on our texture suppression loss to control the areas where we want to smooth and suppress the texture.

\textbf{Masked total variation.} Total variation loss distortion is usually used to reduce noise and artifacts in the generated results. In order to adapt to our color transfer task, we apply a $EdgeMask$ to it to force it to only suppress the texture of the smooth area of the content image. The calculation process of $EdgeMask$ can be defined as:
\begin{equation}
	\begin{aligned}
		\label{eq5}
		\mathcal{L}_{mtv} = \sum_{x=1}^{H}\sum_{y=1}^{W-1}(cs_{c_{x,y+1}}-cs_{c_{x,y}})^{2}\cdot EdgeMask_{x,y+1}\\
		+\sum_{x=1}^{H-1}\sum_{y=1}^{W}(cs_{c_{x+1,y}}-cs_{c_{x,y}})^{2}\cdot EdgeMask_{x+1,y},
	\end{aligned}
\end{equation}
where the $cs_{c}$ is the color transfer result, and H and W are their height and width.
\label{sec:mtv}
\section{Analysis}
\subsection{Texture Suppression Differentiated Performance}
Although the masked total variation loss has been suppressed as much as possible to suppress the texture representation in Gram, as shown in Fig.\ref{guide}(a), subtle texture representations have still been generated. However, there are almost no subtle texture representations in Fig.\ref{guide}(b), and based on this differentiated representation, we attempt to analyze the reasons for its occurrence. We enlarged and compared two content images and found that there were many discontinuous artifacts and noise in the sky part of the airplane image, while the sky of the house image was in a relatively continuous state. Therefore, we assume that the differentiated representation of this texture generation is based on the continuity of the input content image.
\label{sec:tsdp}

\subsection{Input Smoothing}
To verify the impact of image continuity on texture generation, we smoothed the input image. In order to perform smoothing operations without affecting the content structure information as much as possible, we choose to use a guided filtering method, which can maintain edge information while smoothing non edge areas. As shown in Fig.\ref{guide}(c), after smoothing the input image, the subtle texture representations in the image almost completely disappear.

\subsection{Feature Visualization Analysis}
As shown in Fig.\ref{vis}, first we observe the feature maps without $\mathcal{L}_{mtv}$, and we can see that almost all feature maps in Conv1 have subtle noise features. Zoom in on the red box areas of the second and seventh feature maps, where noise features are particularly prominent. And as we go through more convolutions, we see that almost all the channels at Conv3 are covered with obvious linear textures. Looking at the feature maps with $\mathcal{L}_{mtv}$, we can clearly see that both the noise in Conv1 and the number and intensity of feature maps with line structure in Conv3 have significantly decreased. Based on the above observations, we believe that the evolution process from noise to linear texture in the feature map seems to be the step of texture feature generation under this framework, and we verify that $\mathcal{L}_{mtv}$ does have a inhibitory effect on texture generation but cannot be eradicated.

Finally, we observed the feature maps after guided filtering of the content images and clearly found that the noise and texture structure in Conv1 and Conv3 had almost disappeared, confirming our hypothesis in \ref{sec:tsdp}. In the shallow structure of our CTDP framework, we believe that the presentation of this final texture structure needs to start with the first convolution generating noise, and then continuously spread the noise to each subsequent channel through convolution operations, generating a more structured texture with higher receptive fields in this process.
\subsection{Noise addition}
\begin{figure}[t]\centering
	\subfigure[w/o guide filtering]{\includegraphics[width=0.156\textwidth]{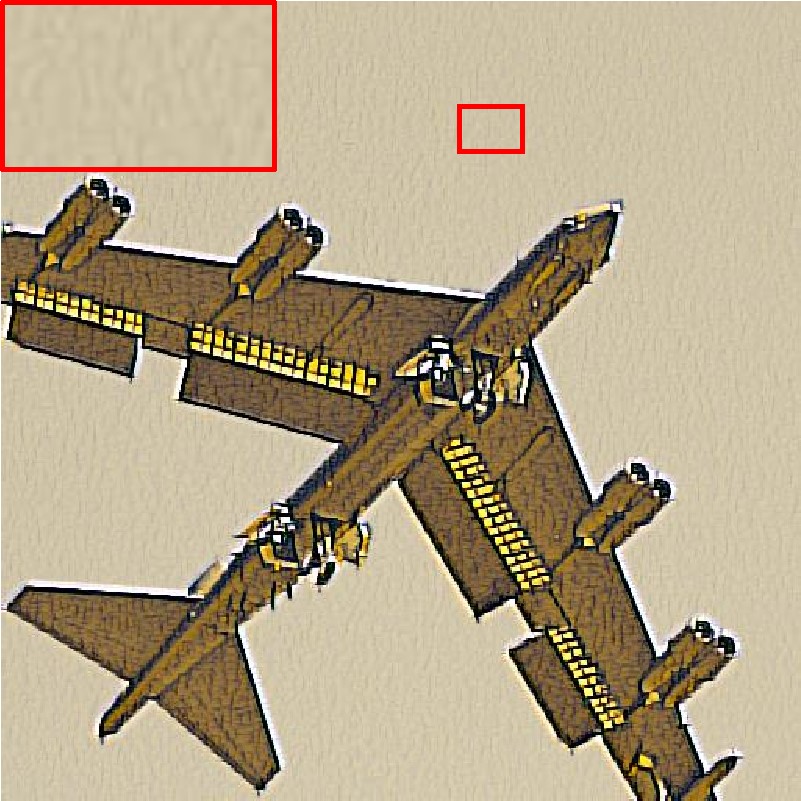}}
	\subfigure[w/o guide filtering]{\includegraphics[width=0.156\textwidth]{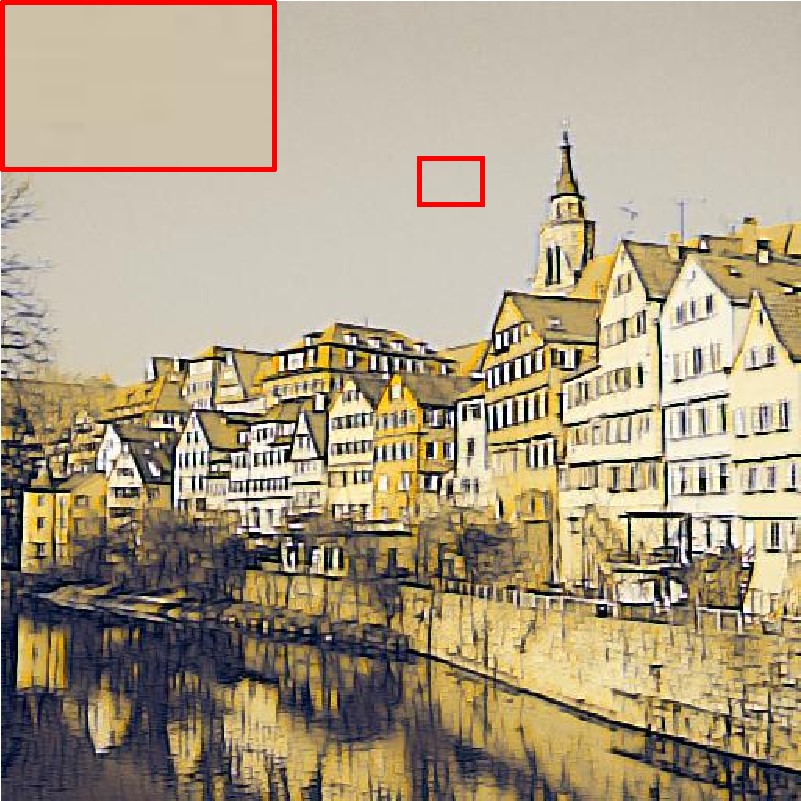}}
	\subfigure[w/ guide filtering]{\includegraphics[width=0.156\textwidth]{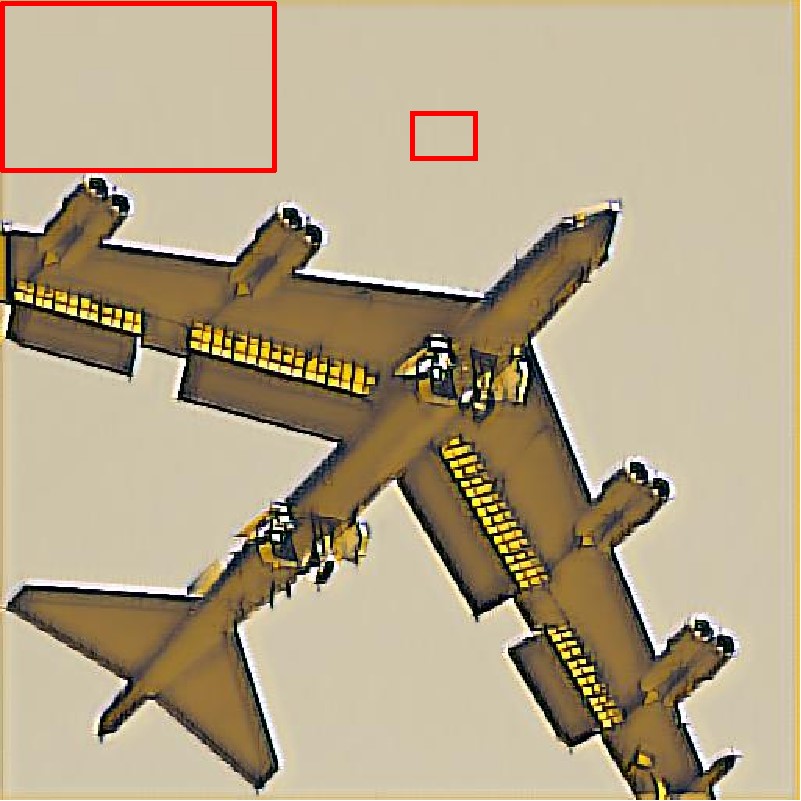}}
	\caption{Texture Suppression differentiated performance.}
	\label{guide}
\end{figure}

\begin{figure}[t]\centering
	
	\subfigure[w/ guide filtering]{\includegraphics[width=0.156\textwidth]{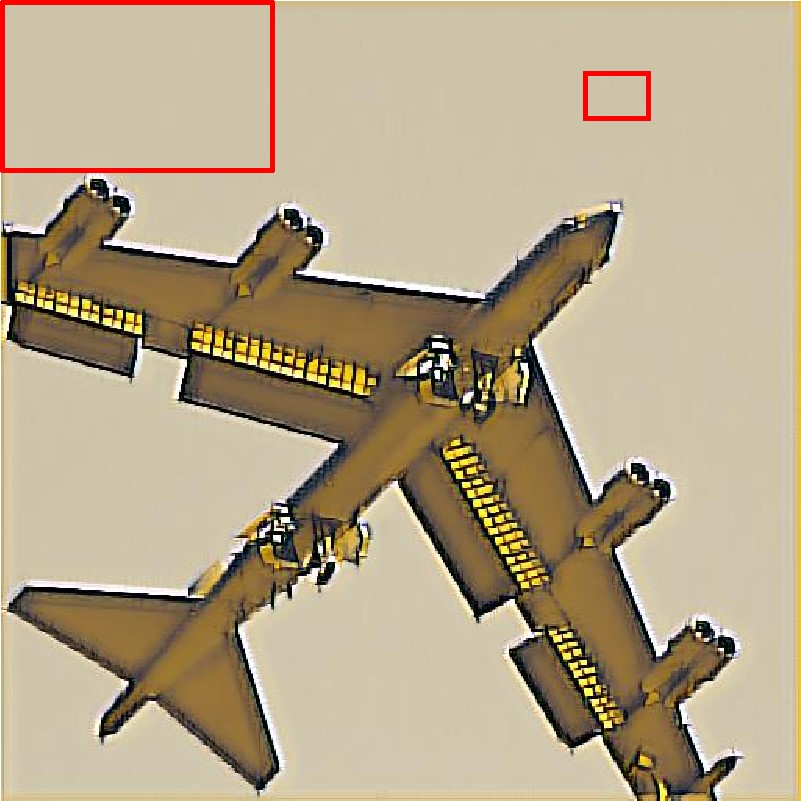}}
	\subfigure[two channels with noise]{\includegraphics[width=0.156\textwidth]{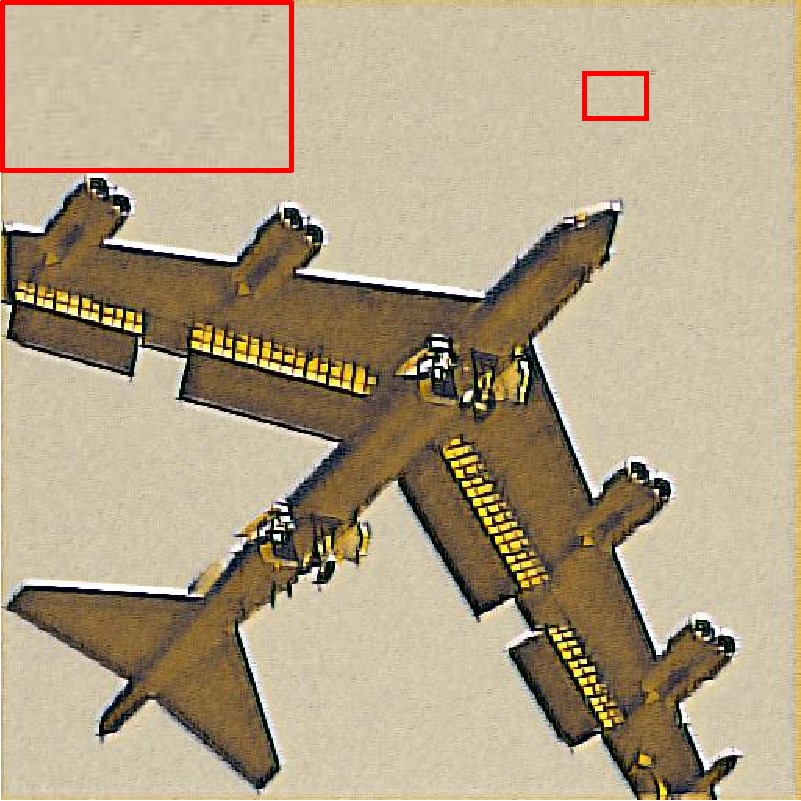}}
	\subfigure[all channels with noise]{\includegraphics[width=0.156\textwidth]{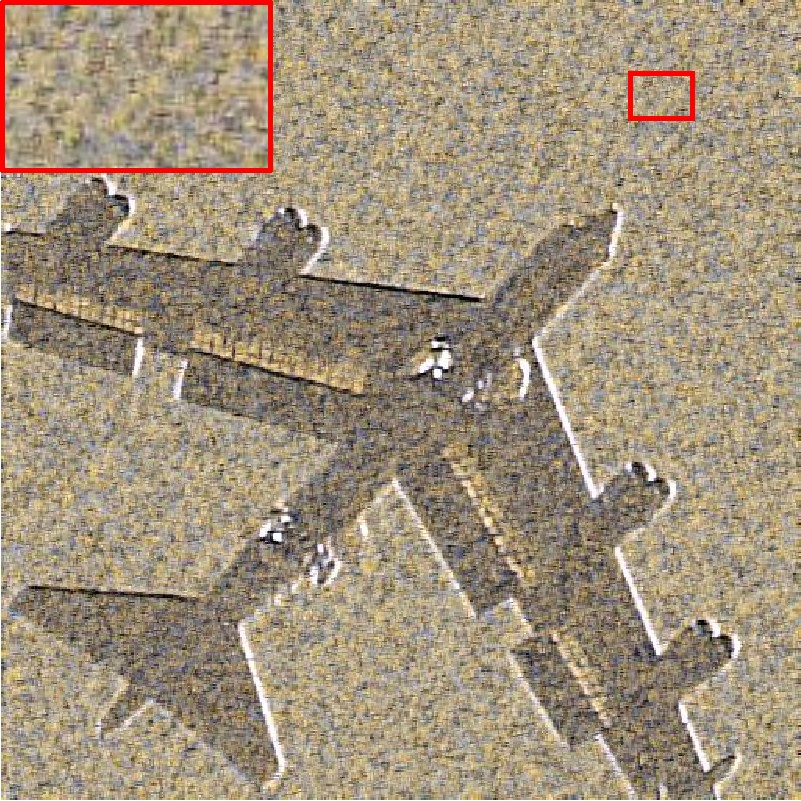}}
	\caption{Adding noise to the feature map for smoothing operations.}
	\label{noise}
\end{figure}

After clarifying the mechanism of CTDP texture generation and its performance in feature maps, we attempted to directly manipulate the feature maps to affect texture generation. As shown in Fig.\ref{noise}(a), we take the output after the smoothing operation as the initial result and observe the effect of adding noise (generating noise from the standard normal distribution) to its smooth feature map. Firstly, we performed noise adding operations on the first and eight feature maps of the guided filtering feature map in Fig.\ref{vis}, and the output results in Fig.\ref{noise}(b) were similar to those in Fig.\ref{guide}(a). Subsequently, we added noise to all channels to produce the effect shown in Fig.\ref{noise}(c), resulting in various complex texture and color block structures. This experiment once again verified that the noise in the first layer feature map is the starting point of texture generation, and the noise in different channels should control different coupled color and texture features.

\section{Experiments}
\subsection{Implementation Details}
We used MS-COO (\cite{coco}) as the content image and extracted style images from Wikiart (\cite{wiki}) to train our CTDP model. In equation.\ref{zongloss}, the values of $\lambda_{bc}$,  $\lambda_{bs}$, $\lambda_{adv}$, $\lambda_{mtv}$ and $\lambda_{fdc}$ are set to 1e0, 1e5, 1e0, 2e-5 and 1e0, respectively. We used the Adam (\cite{adam}) optimizer with a learning rate of 0.001. During the training process, first adjust the size of the content image to 512, and then randomly crop it to 256 $\times$ 256 pixels for enhancement. Use similar methods to process style images, but all images in a batch are randomly cropped from the same style image. It is worth noting that since our CTDP is fully convolutional, it can process input images of any size during testing. We conducted all experiments on RTX 3090 GPU.
\subsection{Comparisons with Prior Arts}
Due to our model's ability to quickly generate color and texture transfer results simultaneously, we compared our CTDP with state-of-the-art color transfer models and texture transfer models (arbitrary style transfer). In the comparison scheme, we directly ran the code with the default settings published by the author.

\begin{figure*}[t]
	\centering
	\includegraphics[width=\textwidth,height=\textheight,keepaspectratio]{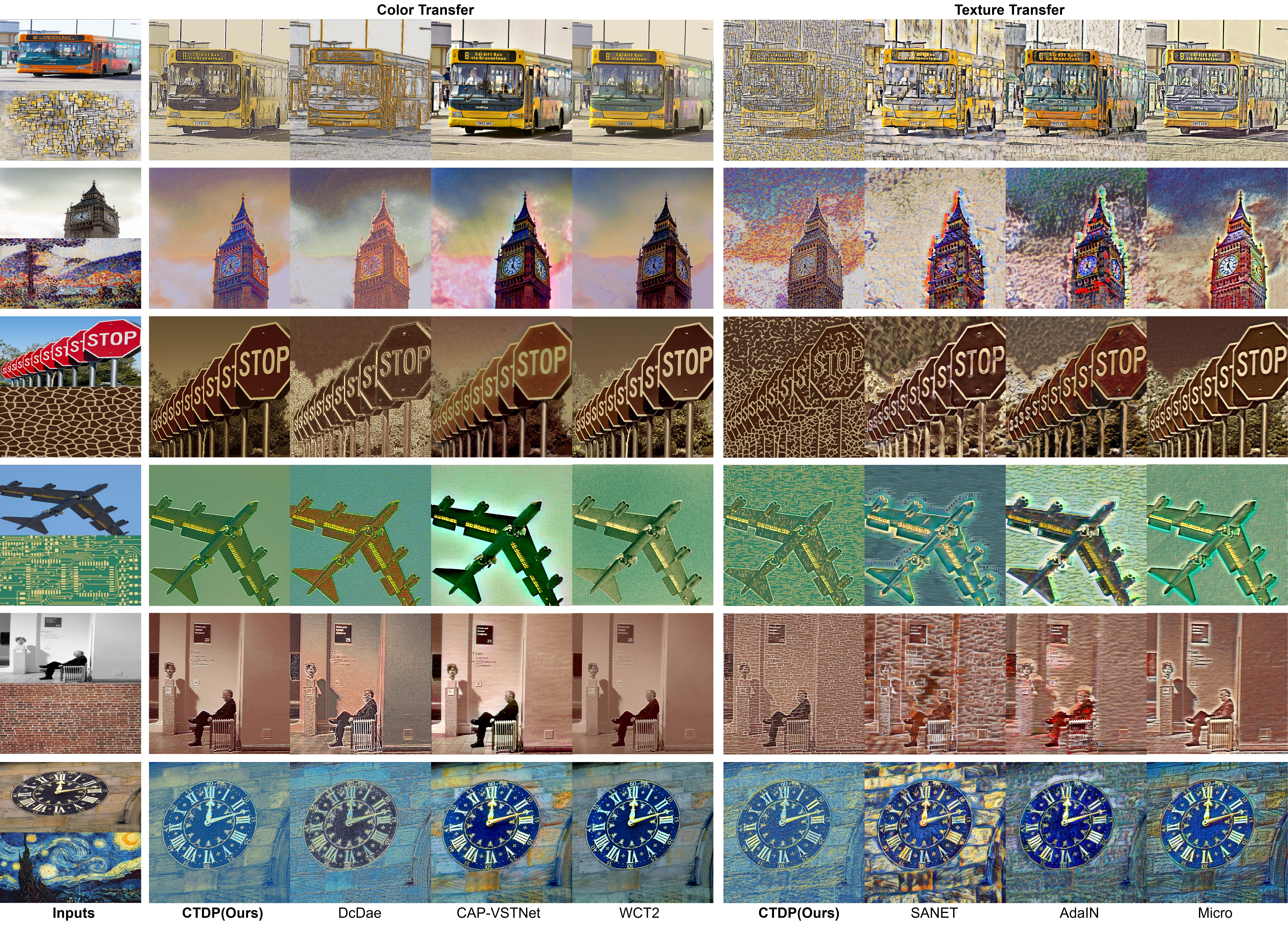}
	\caption{\textbf{Quantitative Comparison} with the state-of-the-art color and texture transfer methods using 1024 resolution input images. Due to the selection of many challenging style images with complex texture structures, it is best to zoom in to better observe artifact suppression and texture structure transfer.}
	\label{duibi}
\end{figure*}
\begin{figure}[t]\centering
	
	\subfigure[Full Model]{\includegraphics[width=0.156\textwidth]{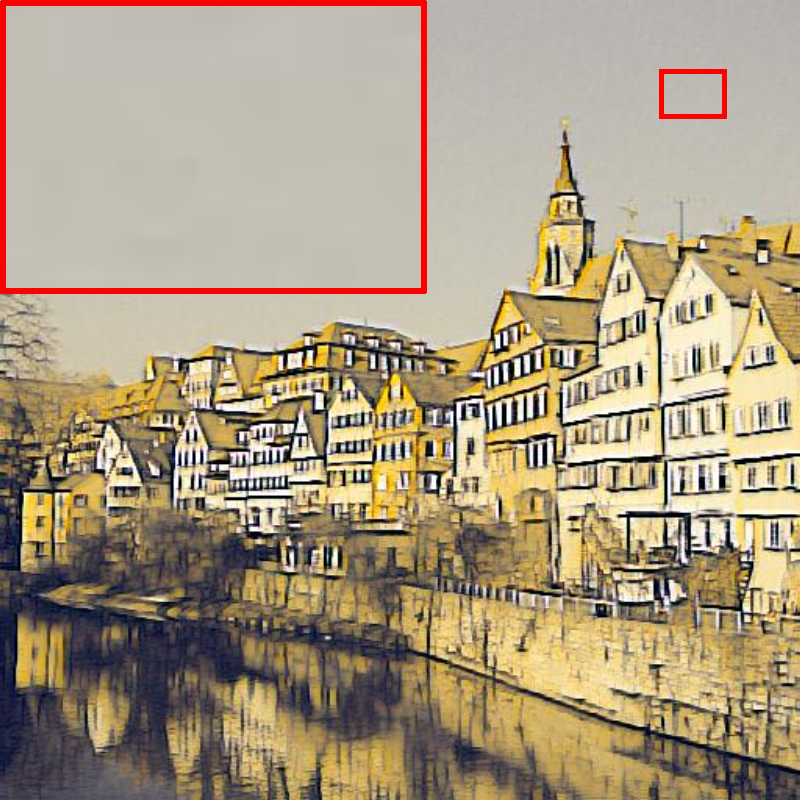}}
	\subfigure[w/o $\mathcal{L}_{bs}$]{\includegraphics[width=0.156\textwidth]{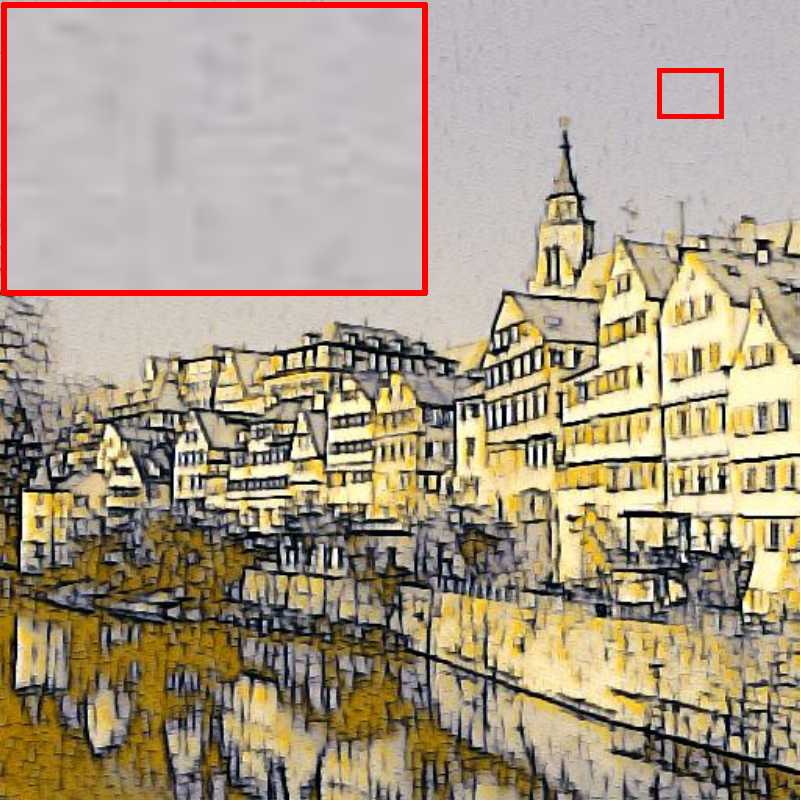}}
	\subfigure[w/o $\mathcal{L}_{mtv}$]{\includegraphics[width=0.156\textwidth]{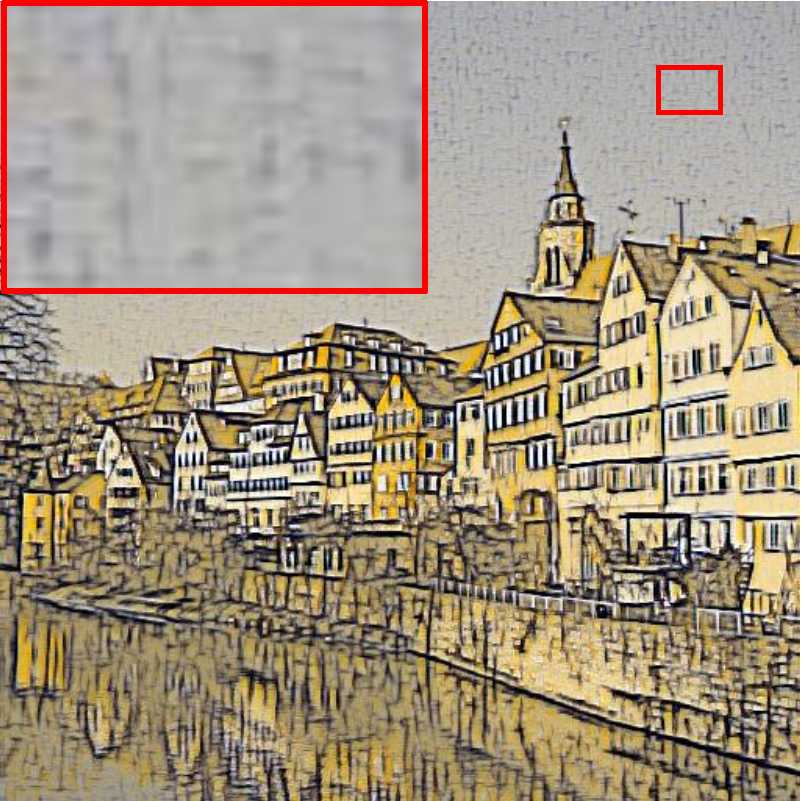}}
	\caption{\textbf{Ablation study} of branch style loss and masked total variation loss  to evaluate their effectiveness in suppressing texture and artifacts in color transfer tasks.}
	\label{A}
\end{figure}

\subsubsection{Qualitative Comparison}
The qualitative comparison results of different color transfer and texture transfer methods are shown on the left and right sides of Fig.\ref{duibi}.

\textbf{Color Transfer.} Firstly, compared with our previous work DcDae (\cite{DcDae}), the direct output of its intermediate features can serve as an approximate result of color transfer tasks, but there may be color mismatches and many obvious texture representations. By comparison, our CTDP has greatly improved the color transfer effect.

CAP-VSTNet (\cite{cap}) and WCT2 (\cite{photowct2}) both exhibit noise and duplicate local feature blocks when facing reference images with rich textures (such as the background parts in the third and fourth rows). Both methods have the problem of content color leakage (such as varying degrees of color leakage in the first row of the vehicle body), and the color leakage in CAP-VSTNet is particularly severe (such as billboards in the third row, aircraft fuselage in the fourth row, and background bricks in the sixth row). Additionally, CAP-VSTNet also produces false effects similar to halos at semantic edges (such as buildings in the second and fourth rows and aircraft contours).

In contrast, our CTDP achieves very advanced color transfer effects. We have done a good job in global color distribution matching, avoiding color leakage issues, and suppressing texture structures.

\textbf{Texture Transfer.}
CTDP has not made any further improvements in texture transfer tasks compared to its previous DcDae, and continues to maintain state-of-the-art texture transfer effects.

Adain (\cite{adain}) has serious color leakage issues (such as first row body, third row billboard, fourth row body), while Micro (\cite{microast}) also has a few color leakage issues (such as first row body). The three methods did not perform well in texture transfer tasks and did not match the texture structure well. The SANET (\cite{SANet}) and Adain methods have greatly distorted the semantics of the content (such as the characters on the first lane head and the characters on the fifth line wall).

In contrast, our CTDP achieves state-of-the-art texture transfer effects. We have done a good job in global color, texture structure distribution matching, content semantic preservation, and avoiding color leakage issues, especially in texture structure transfer.

\subsection{Ablation Study}
The ablation experiment for the additional loss term is shown in Fig.\ref{A}. The case of directly calculating the Gram matching 4 layers in shallow layers instead of the two layers in $\mathcal{L}_{bs}$, as shown in Fig.\ref{A}(b). The texture representation of the reference style shows slight leakage (zooming in on the fine texture features in the red box), and the overall transfer effect is extremely disharmonious due to the small receptive field of the model but the large receptive field of Gram's calculation. This indicates that $\mathcal{L}_{bs}$ is more suitable for guiding color transfer branching. (2) Without $\mathcal{L}_{mtv}$, as shown in the red box in Fig.\ref{A}(c), the texture representation becomes very obvious. This indicates that $\mathcal{L}_{mtv}$ plays a crucial role in suppressing textures. (3) Without $\mathcal{L}_{fdc}$, as shown in Fig.\ref{fdc}(b), the decoding of shallow features through the fusion decoder is still a side effect product of style transfer. This indicates that $\mathcal{L}_{fdc}$ ensures the decoding consistency of shallow features between shallow and fused decoders.
\section{Conclusion}
In this article, we propose a dual pipeline lightweight framework called CTDP. For the first time, our dual channels can simultaneously generate color and texture transfer results corresponding to style images, and the weighted fusion of dual branch features achieves the effect of adding texture features with controllable intensity from color transfer results for the first time. In addition, mtv loss was designed to suppress texture information in the model matching Gram matrix, and it was found that smoothing the input in our framework can almost completely eliminate texture features. A large number of experiments have proven the effectiveness of this method. Compared to the current level of technology, our CTDP is the first model that can simultaneously achieve color and texture transfer. It not only produces visually superior results in both migration tasks, but also has a color migration branch model size as low as 20k.

\bibliographystyle{elsarticle-harv}
\bibliography{ref}

\begin{thebibliography}{43}
\expandafter\ifx\csname natexlab\endcsname\relax\def\natexlab#1{#1}\fi
\providecommand{\url}[1]{\texttt{#1}}
\providecommand{\href}[2]{#2}
\providecommand{\path}[1]{#1}
\providecommand{\DOIprefix}{doi:}
\providecommand{\ArXivprefix}{arXiv:}
\providecommand{\URLprefix}{URL: }
\providecommand{\Pubmedprefix}{pmid:}
\providecommand{\doi}[1]{\href{http://dx.doi.org/#1}{\path{#1}}}
\providecommand{\Pubmed}[1]{\href{pmid:#1}{\path{#1}}}
\providecommand{\bibinfo}[2]{#2}
\ifx\xfnm\relax \def\xfnm[#1]{\unskip,\space#1}\fi
\bibitem[{Champandard(2016)}]{cham}
\bibinfo{author}{Champandard, A.J.}, \bibinfo{year}{2016}.
\newblock \bibinfo{title}{Semantic style transfer and turning two-bit doodles
  into fine artworks}.
\newblock \bibinfo{journal}{arXiv preprint arXiv:1603.01768} .
\bibitem[{Chen et~al.(2017)Chen, Yuan, Liao, Yu and Hua}]{3}
\bibinfo{author}{Chen, D.}, \bibinfo{author}{Yuan, L.}, \bibinfo{author}{Liao,
  J.}, \bibinfo{author}{Yu, N.}, \bibinfo{author}{Hua, G.},
  \bibinfo{year}{2017}.
\newblock \bibinfo{title}{Stylebank: An explicit representation for neural
  image style transfer}, in: \bibinfo{booktitle}{Proceedings of the IEEE
  conference on computer vision and pattern recognition}, pp.
  \bibinfo{pages}{1897--1906}.
\bibitem[{Chen et~al.(2021)Chen, Zhao, Zhang, Wang, Zuo, Li, Xing and
  Lu}]{chen}
\bibinfo{author}{Chen, H.}, \bibinfo{author}{Zhao, L.}, \bibinfo{author}{Zhang,
  H.}, \bibinfo{author}{Wang, Z.}, \bibinfo{author}{Zuo, Z.},
  \bibinfo{author}{Li, A.}, \bibinfo{author}{Xing, W.}, \bibinfo{author}{Lu,
  D.}, \bibinfo{year}{2021}.
\newblock \bibinfo{title}{Diverse image style transfer via invertible
  cross-space mapping}, in: \bibinfo{booktitle}{2021 IEEE/CVF International
  Conference on Computer Vision (ICCV)}, \bibinfo{organization}{IEEE Computer
  Society}. pp. \bibinfo{pages}{14860--14869}.
\bibitem[{Chiu and Gurari(2022)}]{photowct2}
\bibinfo{author}{Chiu, T.Y.}, \bibinfo{author}{Gurari, D.},
  \bibinfo{year}{2022}.
\newblock \bibinfo{title}{Photowct2: Compact autoencoder for photorealistic
  style transfer resulting from blockwise training and skip connections of
  high-frequency residuals}, in: \bibinfo{booktitle}{Proceedings of the
  IEEE/CVF Winter Conference on Applications of Computer Vision}, pp.
  \bibinfo{pages}{2868--2877}.
\bibitem[{Dumoulin et~al.(2016)Dumoulin, Shlens and Kudlur}]{7}
\bibinfo{author}{Dumoulin, V.}, \bibinfo{author}{Shlens, J.},
  \bibinfo{author}{Kudlur, M.}, \bibinfo{year}{2016}.
\newblock \bibinfo{title}{A learned representation for artistic style}.
\newblock \bibinfo{journal}{arXiv preprint arXiv:1610.07629} .
\bibitem[{Gatys et~al.(2016)Gatys, Ecker and Bethge}]{gatys}
\bibinfo{author}{Gatys, L.A.}, \bibinfo{author}{Ecker, A.S.},
  \bibinfo{author}{Bethge, M.}, \bibinfo{year}{2016}.
\newblock \bibinfo{title}{Image style transfer using convolutional neural
  networks}, in: \bibinfo{booktitle}{Proceedings of the IEEE conference on
  computer vision and pattern recognition}, pp. \bibinfo{pages}{2414--2423}.
\bibitem[{Gu et~al.(2018)Gu, Chen, Liao and Yuan}]{10}
\bibinfo{author}{Gu, S.}, \bibinfo{author}{Chen, C.}, \bibinfo{author}{Liao,
  J.}, \bibinfo{author}{Yuan, L.}, \bibinfo{year}{2018}.
\newblock \bibinfo{title}{Arbitrary style transfer with deep feature
  reshuffle}, in: \bibinfo{booktitle}{Proceedings of the IEEE Conference on
  Computer Vision and Pattern Recognition}, pp. \bibinfo{pages}{8222--8231}.
\bibitem[{Howard et~al.(2017)Howard, Zhu, Chen, Kalenichenko, Wang, Weyand,
  Andreetto and Adam}]{mobilenets}
\bibinfo{author}{Howard, A.G.}, \bibinfo{author}{Zhu, M.},
  \bibinfo{author}{Chen, B.}, \bibinfo{author}{Kalenichenko, D.},
  \bibinfo{author}{Wang, W.}, \bibinfo{author}{Weyand, T.},
  \bibinfo{author}{Andreetto, M.}, \bibinfo{author}{Adam, H.},
  \bibinfo{year}{2017}.
\newblock \bibinfo{title}{Mobilenets: Efficient convolutional neural networks
  for mobile vision applications}.
\newblock \bibinfo{journal}{arXiv preprint arXiv:1704.04861} .
\bibitem[{Huang and Belongie(2017)}]{adain}
\bibinfo{author}{Huang, X.}, \bibinfo{author}{Belongie, S.},
  \bibinfo{year}{2017}.
\newblock \bibinfo{title}{Arbitrary style transfer in real-time with adaptive
  instance normalization}, in: \bibinfo{booktitle}{Proceedings of the IEEE
  international conference on computer vision}, pp.
  \bibinfo{pages}{1501--1510}.
\bibitem[{Jing et~al.(2020)Jing, Liu, Ding, Wang, Ding, Song and
  Wen}]{DynamicIN}
\bibinfo{author}{Jing, Y.}, \bibinfo{author}{Liu, X.}, \bibinfo{author}{Ding,
  Y.}, \bibinfo{author}{Wang, X.}, \bibinfo{author}{Ding, E.},
  \bibinfo{author}{Song, M.}, \bibinfo{author}{Wen, S.}, \bibinfo{year}{2020}.
\newblock \bibinfo{title}{Dynamic instance normalization for arbitrary style
  transfer}, in: \bibinfo{booktitle}{Proceedings of the AAAI Conference on
  Artificial Intelligence}, pp. \bibinfo{pages}{4369--4376}.
\bibitem[{Jing et~al.(2018)Jing, Liu, Yang, Feng, Yu, Tao and Song}]{stroke}
\bibinfo{author}{Jing, Y.}, \bibinfo{author}{Liu, Y.}, \bibinfo{author}{Yang,
  Y.}, \bibinfo{author}{Feng, Z.}, \bibinfo{author}{Yu, Y.},
  \bibinfo{author}{Tao, D.}, \bibinfo{author}{Song, M.}, \bibinfo{year}{2018}.
\newblock \bibinfo{title}{Stroke controllable fast style transfer with adaptive
  receptive fields}, in: \bibinfo{booktitle}{Proceedings of the European
  Conference on Computer Vision (ECCV)}, pp. \bibinfo{pages}{238--254}.
\bibitem[{Johnson et~al.(2016)Johnson, Alahi and Fei-Fei}]{Perceptual}
\bibinfo{author}{Johnson, J.}, \bibinfo{author}{Alahi, A.},
  \bibinfo{author}{Fei-Fei, L.}, \bibinfo{year}{2016}.
\newblock \bibinfo{title}{Perceptual losses for real-time style transfer and
  super-resolution}, in: \bibinfo{booktitle}{Computer Vision--ECCV 2016: 14th
  European Conference, Amsterdam, The Netherlands, October 11-14, 2016,
  Proceedings, Part II 14}, \bibinfo{organization}{Springer}. pp.
  \bibinfo{pages}{694--711}.
\bibitem[{Kingma and Ba(2014)}]{adam}
\bibinfo{author}{Kingma, D.P.}, \bibinfo{author}{Ba, J.}, \bibinfo{year}{2014}.
\newblock \bibinfo{title}{Adam: A method for stochastic optimization}.
\newblock \bibinfo{journal}{arXiv preprint arXiv:1412.6980} .
\bibitem[{Kolkin et~al.(2019)Kolkin, Salavon and Shakhnarovich}]{15}
\bibinfo{author}{Kolkin, N.}, \bibinfo{author}{Salavon, J.},
  \bibinfo{author}{Shakhnarovich, G.}, \bibinfo{year}{2019}.
\newblock \bibinfo{title}{Style transfer by relaxed optimal transport and
  self-similarity}, in: \bibinfo{booktitle}{Proceedings of the IEEE/CVF
  Conference on Computer Vision and Pattern Recognition}, pp.
  \bibinfo{pages}{10051--10060}.
\bibitem[{Li and Wand(2016a)}]{17}
\bibinfo{author}{Li, C.}, \bibinfo{author}{Wand, M.}, \bibinfo{year}{2016}a.
\newblock \bibinfo{title}{Combining markov random fields and convolutional
  neural networks for image synthesis}, in: \bibinfo{booktitle}{Proceedings of
  the IEEE conference on computer vision and pattern recognition}, pp.
  \bibinfo{pages}{2479--2486}.
\bibitem[{Li and Wand(2016b)}]{18}
\bibinfo{author}{Li, C.}, \bibinfo{author}{Wand, M.}, \bibinfo{year}{2016}b.
\newblock \bibinfo{title}{Precomputed real-time texture synthesis with
  markovian generative adversarial networks}, in: \bibinfo{booktitle}{Computer
  Vision--ECCV 2016: 14th European Conference, Amsterdam, The Netherlands,
  October 11-14, 2016, Proceedings, Part III 14},
  \bibinfo{organization}{Springer}. pp. \bibinfo{pages}{702--716}.
\bibitem[{Li et~al.(2017a)Li, Fang, Yang, Wang, Lu and Yang}]{20}
\bibinfo{author}{Li, Y.}, \bibinfo{author}{Fang, C.}, \bibinfo{author}{Yang,
  J.}, \bibinfo{author}{Wang, Z.}, \bibinfo{author}{Lu, X.},
  \bibinfo{author}{Yang, M.H.}, \bibinfo{year}{2017}a.
\newblock \bibinfo{title}{Diversified texture synthesis with feed-forward
  networks}, in: \bibinfo{booktitle}{Proceedings of the IEEE conference on
  computer vision and pattern recognition}, pp. \bibinfo{pages}{3920--3928}.
\bibitem[{Li et~al.(2017b)Li, Fang, Yang, Wang, Lu and Yang}]{21}
\bibinfo{author}{Li, Y.}, \bibinfo{author}{Fang, C.}, \bibinfo{author}{Yang,
  J.}, \bibinfo{author}{Wang, Z.}, \bibinfo{author}{Lu, X.},
  \bibinfo{author}{Yang, M.H.}, \bibinfo{year}{2017}b.
\newblock \bibinfo{title}{Universal style transfer via feature transforms}.
\newblock \bibinfo{journal}{Advances in neural information processing systems}
  \bibinfo{volume}{30}.
\bibitem[{Li et~al.(2018)Li, Liu, Li, Yang and Kautz}]{closed}
\bibinfo{author}{Li, Y.}, \bibinfo{author}{Liu, M.Y.}, \bibinfo{author}{Li,
  X.}, \bibinfo{author}{Yang, M.H.}, \bibinfo{author}{Kautz, J.},
  \bibinfo{year}{2018}.
\newblock \bibinfo{title}{A closed-form solution to photorealistic image
  stylization}, in: \bibinfo{booktitle}{Proceedings of the European conference
  on computer vision (ECCV)}, pp. \bibinfo{pages}{453--468}.
\bibitem[{Li et~al.(2017c)Li, Wang, Liu and Hou}]{demystifying}
\bibinfo{author}{Li, Y.}, \bibinfo{author}{Wang, N.}, \bibinfo{author}{Liu,
  J.}, \bibinfo{author}{Hou, X.}, \bibinfo{year}{2017}c.
\newblock \bibinfo{title}{Demystifying neural style transfer}.
\newblock \bibinfo{journal}{arXiv preprint arXiv:1701.01036} .
\bibitem[{Lin et~al.(2014)Lin, Maire, Belongie, Hays, Perona, Ramanan,
  Doll{\'a}r and Zitnick}]{coco}
\bibinfo{author}{Lin, T.Y.}, \bibinfo{author}{Maire, M.},
  \bibinfo{author}{Belongie, S.}, \bibinfo{author}{Hays, J.},
  \bibinfo{author}{Perona, P.}, \bibinfo{author}{Ramanan, D.},
  \bibinfo{author}{Doll{\'a}r, P.}, \bibinfo{author}{Zitnick, C.L.},
  \bibinfo{year}{2014}.
\newblock \bibinfo{title}{Microsoft coco: Common objects in context}, in:
  \bibinfo{booktitle}{Computer Vision--ECCV 2014: 13th European Conference,
  Zurich, Switzerland, September 6-12, 2014, Proceedings, Part V 13},
  \bibinfo{organization}{Springer}. pp. \bibinfo{pages}{740--755}.
\bibitem[{Luan et~al.(2017)Luan, Paris, Shechtman and Bala}]{deep}
\bibinfo{author}{Luan, F.}, \bibinfo{author}{Paris, S.},
  \bibinfo{author}{Shechtman, E.}, \bibinfo{author}{Bala, K.},
  \bibinfo{year}{2017}.
\newblock \bibinfo{title}{Deep photo style transfer}, in:
  \bibinfo{booktitle}{Proceedings of the IEEE conference on computer vision and
  pattern recognition}, pp. \bibinfo{pages}{4990--4998}.
\bibitem[{Park and Lee(2019)}]{SANet}
\bibinfo{author}{Park, D.Y.}, \bibinfo{author}{Lee, K.H.},
  \bibinfo{year}{2019}.
\newblock \bibinfo{title}{Arbitrary style transfer with style-attentional
  networks}, in: \bibinfo{booktitle}{proceedings of the IEEE/CVF conference on
  computer vision and pattern recognition}, pp. \bibinfo{pages}{5880--5888}.
\bibitem[{Phillips and Mackintosh(2011)}]{wiki}
\bibinfo{author}{Phillips, F.}, \bibinfo{author}{Mackintosh, B.},
  \bibinfo{year}{2011}.
\newblock \bibinfo{title}{Wiki art gallery, inc.: A case for critical
  thinking}.
\newblock \bibinfo{journal}{Issues in Accounting Education}
  \bibinfo{volume}{26}, \bibinfo{pages}{593--608}.
\bibitem[{Pitie et~al.(2005)Pitie, Kokaram and Dahyot}]{pitie2005n}
\bibinfo{author}{Pitie, F.}, \bibinfo{author}{Kokaram, A.C.},
  \bibinfo{author}{Dahyot, R.}, \bibinfo{year}{2005}.
\newblock \bibinfo{title}{N-dimensional probability density function transfer
  and its application to color transfer}, in: \bibinfo{booktitle}{Tenth IEEE
  International Conference on Computer Vision (ICCV'05) Volume 1},
  \bibinfo{organization}{IEEE}. pp. \bibinfo{pages}{1434--1439}.
\bibitem[{Piti{\'e} et~al.(2007)Piti{\'e}, Kokaram and
  Dahyot}]{pitie2007automated}
\bibinfo{author}{Piti{\'e}, F.}, \bibinfo{author}{Kokaram, A.C.},
  \bibinfo{author}{Dahyot, R.}, \bibinfo{year}{2007}.
\newblock \bibinfo{title}{Automated colour grading using colour distribution
  transfer}.
\newblock \bibinfo{journal}{Computer Vision and Image Understanding}
  \bibinfo{volume}{107}, \bibinfo{pages}{123--137}.
\bibitem[{Reinhard et~al.(2001)Reinhard, Adhikhmin, Gooch and
  Shirley}]{reinhard2001color}
\bibinfo{author}{Reinhard, E.}, \bibinfo{author}{Adhikhmin, M.},
  \bibinfo{author}{Gooch, B.}, \bibinfo{author}{Shirley, P.},
  \bibinfo{year}{2001}.
\newblock \bibinfo{title}{Color transfer between images}.
\newblock \bibinfo{journal}{IEEE Computer graphics and applications}
  \bibinfo{volume}{21}, \bibinfo{pages}{34--41}.
\bibitem[{Risser et~al.(2017)Risser, Wilmot and Barnes}]{27}
\bibinfo{author}{Risser, E.}, \bibinfo{author}{Wilmot, P.},
  \bibinfo{author}{Barnes, C.}, \bibinfo{year}{2017}.
\newblock \bibinfo{title}{Stable and controllable neural texture synthesis and
  style transfer using histogram losses}.
\newblock \bibinfo{journal}{arXiv preprint arXiv:1701.08893} .
\bibitem[{Sengupta et~al.(2019)Sengupta, Ye, Wang, Liu and Roy}]{vgg}
\bibinfo{author}{Sengupta, A.}, \bibinfo{author}{Ye, Y.},
  \bibinfo{author}{Wang, R.}, \bibinfo{author}{Liu, C.}, \bibinfo{author}{Roy,
  K.}, \bibinfo{year}{2019}.
\newblock \bibinfo{title}{Going deeper in spiking neural networks: Vgg and
  residual architectures}.
\newblock \bibinfo{journal}{Frontiers in neuroscience} \bibinfo{volume}{13},
  \bibinfo{pages}{95}.
\bibitem[{Shen et~al.(2018)Shen, Yan and Zeng}]{meta}
\bibinfo{author}{Shen, F.}, \bibinfo{author}{Yan, S.}, \bibinfo{author}{Zeng,
  G.}, \bibinfo{year}{2018}.
\newblock \bibinfo{title}{Neural style transfer via meta networks}, in:
  \bibinfo{booktitle}{Proceedings of the IEEE Conference on Computer Vision and
  Pattern Recognition}, pp. \bibinfo{pages}{8061--8069}.
\bibitem[{ShiQi~Jiang(2023)}]{DcDae}
\bibinfo{author}{ShiQi~Jiang, JunJie~Kang, Y.L.}, \bibinfo{year}{2023}.
\newblock \bibinfo{title}{Degree-controllable lightweight fast style transfer
  with detail attention-enhanced}.
\newblock \bibinfo{journal}{arXiv preprint arXiv:2306.16846} .
\bibitem[{Ulyanov et~al.(2016a)Ulyanov, Lebedev, Vedaldi and Lempitsky}]{32}
\bibinfo{author}{Ulyanov, D.}, \bibinfo{author}{Lebedev, V.},
  \bibinfo{author}{Vedaldi, A.}, \bibinfo{author}{Lempitsky, V.},
  \bibinfo{year}{2016}a.
\newblock \bibinfo{title}{Texture networks: Feed-forward synthesis of textures
  and stylized images}.
\newblock \bibinfo{journal}{arXiv preprint arXiv:1603.03417} .
\bibitem[{Ulyanov et~al.(2016b)Ulyanov, Vedaldi and Lempitsky}]{33}
\bibinfo{author}{Ulyanov, D.}, \bibinfo{author}{Vedaldi, A.},
  \bibinfo{author}{Lempitsky, V.}, \bibinfo{year}{2016}b.
\newblock \bibinfo{title}{Instance normalization: The missing ingredient for
  fast stylization}.
\newblock \bibinfo{journal}{arXiv preprint arXiv:1607.08022} .
\bibitem[{Wang et~al.(2020)Wang, Li, Wang, Hu and Yang}]{collaborative}
\bibinfo{author}{Wang, H.}, \bibinfo{author}{Li, Y.}, \bibinfo{author}{Wang,
  Y.}, \bibinfo{author}{Hu, H.}, \bibinfo{author}{Yang, M.H.},
  \bibinfo{year}{2020}.
\newblock \bibinfo{title}{Collaborative distillation for ultra-resolution
  universal style transfer}, in: \bibinfo{booktitle}{Proceedings of the
  IEEE/CVF conference on computer vision and pattern recognition}, pp.
  \bibinfo{pages}{1860--1869}.
\bibitem[{Wang et~al.(2017)Wang, Oxholm, Zhang and Wang}]{35}
\bibinfo{author}{Wang, X.}, \bibinfo{author}{Oxholm, G.},
  \bibinfo{author}{Zhang, D.}, \bibinfo{author}{Wang, Y.F.},
  \bibinfo{year}{2017}.
\newblock \bibinfo{title}{Multimodal transfer: A hierarchical deep
  convolutional neural network for fast artistic style transfer}, in:
  \bibinfo{booktitle}{Proceedings of the IEEE conference on computer vision and
  pattern recognition}, pp. \bibinfo{pages}{5239--5247}.
\bibitem[{Wang et~al.(2021)Wang, Zhao, Chen, Zuo, Li, Xing and Lu}]{wang}
\bibinfo{author}{Wang, Z.}, \bibinfo{author}{Zhao, L.}, \bibinfo{author}{Chen,
  H.}, \bibinfo{author}{Zuo, Z.}, \bibinfo{author}{Li, A.},
  \bibinfo{author}{Xing, W.}, \bibinfo{author}{Lu, D.}, \bibinfo{year}{2021}.
\newblock \bibinfo{title}{Divswapper: towards diversified patch-based arbitrary
  style transfer}.
\newblock \bibinfo{journal}{arXiv preprint arXiv:2101.06381} .
\bibitem[{Wang et~al.(2022)Wang, Zhao, Zuo, Li, Chen, Xing and Lu}]{microast}
\bibinfo{author}{Wang, Z.}, \bibinfo{author}{Zhao, L.}, \bibinfo{author}{Zuo,
  Z.}, \bibinfo{author}{Li, A.}, \bibinfo{author}{Chen, H.},
  \bibinfo{author}{Xing, W.}, \bibinfo{author}{Lu, D.}, \bibinfo{year}{2022}.
\newblock \bibinfo{title}{Microast: Towards super-fast ultra-resolution
  arbitrary style transfer}.
\newblock \bibinfo{journal}{arXiv preprint arXiv:2211.15313} .
\bibitem[{Wen et~al.(2023)Wen, Gao and Zou}]{cap}
\bibinfo{author}{Wen, L.}, \bibinfo{author}{Gao, C.}, \bibinfo{author}{Zou,
  C.}, \bibinfo{year}{2023}.
\newblock \bibinfo{title}{Cap-vstnet: Content affinity preserved versatile
  style transfer}, in: \bibinfo{booktitle}{Proceedings of the IEEE/CVF
  Conference on Computer Vision and Pattern Recognition}, pp.
  \bibinfo{pages}{18300--18309}.
\bibitem[{Xie et~al.(2022)Xie, Li, Huang, Fu, Wang and Guo}]{xie}
\bibinfo{author}{Xie, X.}, \bibinfo{author}{Li, Y.}, \bibinfo{author}{Huang,
  H.}, \bibinfo{author}{Fu, H.}, \bibinfo{author}{Wang, W.},
  \bibinfo{author}{Guo, Y.}, \bibinfo{year}{2022}.
\newblock \bibinfo{title}{Artistic style discovery with independent
  components}, in: \bibinfo{booktitle}{Proceedings of the IEEE/CVF Conference
  on Computer Vision and Pattern Recognition}, pp.
  \bibinfo{pages}{19870--19879}.
\bibitem[{Yoo et~al.(2019)Yoo, Uh, Chun, Kang and Ha}]{photorealistic}
\bibinfo{author}{Yoo, J.}, \bibinfo{author}{Uh, Y.}, \bibinfo{author}{Chun,
  S.}, \bibinfo{author}{Kang, B.}, \bibinfo{author}{Ha, J.W.},
  \bibinfo{year}{2019}.
\newblock \bibinfo{title}{Photorealistic style transfer via wavelet
  transforms}, in: \bibinfo{booktitle}{Proceedings of the IEEE/CVF
  International Conference on Computer Vision}, pp.
  \bibinfo{pages}{9036--9045}.
\bibitem[{Zhang et~al.(2019)Zhang, Zhu and Zhu}]{zhang}
\bibinfo{author}{Zhang, C.}, \bibinfo{author}{Zhu, Y.}, \bibinfo{author}{Zhu,
  S.C.}, \bibinfo{year}{2019}.
\newblock \bibinfo{title}{Metastyle: Three-way trade-off among speed,
  flexibility, and quality in neural style transfer}, in:
  \bibinfo{booktitle}{Proceedings of the AAAI Conference on Artificial
  Intelligence}, pp. \bibinfo{pages}{1254--1261}.
\bibitem[{Zhang and Dana(2018a)}]{37}
\bibinfo{author}{Zhang, H.}, \bibinfo{author}{Dana, K.}, \bibinfo{year}{2018}a.
\newblock \bibinfo{title}{Multi-style generative network for real-time
  transfer}, in: \bibinfo{booktitle}{Proceedings of the European Conference on
  Computer Vision (ECCV) Workshops}, pp. \bibinfo{pages}{0--0}.
\bibitem[{Zhang and Dana(2018b)}]{12}
\bibinfo{author}{Zhang, H.}, \bibinfo{author}{Dana, K.}, \bibinfo{year}{2018}b.
\newblock \bibinfo{title}{Multi-style generative network for real-time
  transfer}, in: \bibinfo{booktitle}{Proceedings of the European Conference on
  Computer Vision (ECCV) Workshops}, pp. \bibinfo{pages}{0--0}.

\end{thebibliography}

\end{document}